%% file: main.tex
\documentclass[sigconf]{aamas} 
\usepackage{balance} % for balancing columns on the final page
\usepackage{amsmath}
\usepackage{graphicx}
\usepackage{url}
\usepackage{multirow}
\usepackage{multicol}
\usepackage{hyperref}
\usepackage{diagbox}
\usepackage{enumitem}
\usepackage{indentfirst}

\DeclareMathOperator{\argmax}{\text{argmax}}
\DeclareMathOperator{\argmin}{\text{argmin}}

\setcopyright{ifaamas}
\acmConference[AAMAS '24]{Proc.\@ of the 23rd International Conference
on Autonomous Agents and Multiagent Systems (AAMAS 2024)}{May 6 -- 10, 2024}
{Auckland, New Zealand}{N.~Alechina, V.~Dignum, M.~Dastani, J.S.~Sichman (eds.)}
\copyrightyear{2024}
\acmYear{2024}
\acmDOI{}
\acmPrice{}
\acmISBN{}

\acmSubmissionID{59}

\title{Game-theoretic Objective Space Planning}

\author{Hongrui Zheng}
\affiliation{
  \institution{University of Pennsylvania}
  \city{Philadelphia}
  \country{USA}}
\email{hongruiz@seas.upenn.edu}

\author{Zhijun Zhuang}
\affiliation{
  \institution{University of Pennsylvania}
  \city{Philadelphia}
  \country{USA}}
\email{zhijunz@seas.upenn.edu}

\author{Johannes Betz}
\affiliation{
  \institution{Technical University of Munich}
  \city{Munich}
  \country{Germany}}
\email{johannes.betz@tum.de}

\author{Rahul Mangharam}
\affiliation{
  \institution{University of Pennsylvania}
  \city{Philadelphia}
  \country{USA}}
\email{rahulm@seas.upenn.edu}

\input{abstract.tex}
\keywords{Agent Action Discretization, Game-theoretic Planning, Autonomous Agents, Adversarial Agents, Integrated Planning and Learning, Planning in Latent Space}

\newcommand{\BibTeX}{\rm B\kern-.05em{\sc i\kern-.025em b}\kern-.08em\TeX}

\begin{document}

\maketitle

\input{intro.tex}
\input{method.tex}
\input{experiment.tex}
\input{discussion.tex}

\newpage
\input{impact.tex}
\bibliographystyle{ACM-Reference-Format} 
\bibliography{main}

\newpage
\input{appendix}
\end{document}

%% file: abstract.tex
\begin{abstract}
    Generating competitive strategies and performing continuous motion planning simultaneously in an adversarial setting is a challenging problem. In addition, understanding the intent of other agents is crucial to deploying autonomous systems in adversarial multi-agent environments. Existing approaches either discretize agent action by grouping similar control inputs, sacrificing performance in motion planning, or plan in uninterpretable latent spaces, producing hard-to-understand agent behaviors. Furthermore, the most popular policy optimization frameworks do not recognize the long-term effect of actions and become myopic. This paper proposes an agent action discretization method via abstraction that provides clear intentions of agent actions, an efficient offline pipeline of agent population synthesis, and a planning strategy using counterfactual regret minimization with function approximation. Finally, we experimentally validate our findings on scaled autonomous vehicles in a head-to-head racing setting. We demonstrate that using the proposed framework significantly improves learning, improves the win rate against different opponents, and the improvements can be transferred to unseen opponents in an unseen environment.
\end{abstract}

%% file: intro.tex
\section{Introduction}
Motion planning for autonomous agents in adversarial settings remains a challenging problem, especially for systems with continuous dynamics, where the state, action, and observation spaces are uncountably infinite.
Consider, for example, a head-to-head race between two autonomous race cars. In order to be competitive, an agent not only needs to perform in the highly dynamically challenging task when there are no other agents but also has to adjust their strategy in the presence of another equally competitive agent.

% This problem setting poses two challenges. 
% First, understanding the strategic characteristics of your opponents and creating a continuous motion plan for highly agile agents is challenging. Existing solutions pose significant assumptions about the dynamics, and an optimization had to be set up to solve the game. In contrast, we propose a method that does not pose such assumptions and still creates a continuous motion plan for the agents by planning in a surrogate space.
% Second, since the state space of the actions is continuous, the corresponding game tree will also be infinitely large. Our proposed method provides a tractable way to approximate the outcome of traversing a branch on the game tree, and provide explainable actions for agents participating in the game.

Usually, autonomous agents in these settings are studied as a partially observed Markov decision process (POMDP). A POMDP is represented by the tuple $(\mathcal{S},\mathcal{A},P_{\mathcal{SA}},\mathcal{O},r,\gamma)$ with state space $\mathcal{S}$, action space $\mathcal{A}$, state-action transition probabilities $P_{\mathcal{SA}}$, observation space $\mathcal{O}$, rewards $r:\mathcal{S\times A}\rightarrow \mathbb{R}$ and discount factor $\gamma$. The objective of each agent in the setting is to find the optimal policy $\pi:\mathcal{O}\rightarrow \mathcal{A}$ with parameterization $\theta$ by maximizing the cumulative discounted expected rewards.
% \begin{equation}\label{eq:pomdp_obj}
% \pi^{*} = \underset{\pi}{\argmax} \sum_{t}\gamma^{t}\underset{P_{\mathcal{SA}}}{\mathbb{E}}[r(\pi(\mathcal{O}(t)))]
% \end{equation}
\begin{equation}\label{eq:pomdp_obj}
    \theta^{*} = \underset{\theta}{\argmax} \sum_{t}\gamma^{t}\underset{P_{\mathcal{SA}}}{\mathbb{E}}[r(\pi_{\theta}(\mathcal{O}(t)))]
\end{equation}
However, several challenges arise when the state, action, and observation space of the agents are continuous and uncountably infinite.

\begin{enumerate}[listparindent=2.5em, labelsep=0.5em, itemindent=1.5em, leftmargin=0em]
    \item Continuous action space creates additional complexity when optimizing for a policy. Traditional motion planners and optimal controllers are excellent at dealing with this, but it is challenging to optimize them to perform well in an adversarial setting. The first choice for existing solutions is to model the policy $\pi_{\lambda}(s,a)$ as a distribution with a neural network that predicts the corresponding parameters $\lambda$. The actions are then sampled from the distribution. Alternatively, methods discretize the action space of agents, including binning the control input into intervals or using a bang-bang control. In the first case, bang-bang control-like behavior has been observed in learned policies that use distributions~\cite{huang_learning_2019,novati_remember_2019,thuruthel_model-based_2019}. It has been shown that replacing Gaussian distribution-based policies with bang-bang controllers retains the same level of performance~\cite{seyde_is_2021}. Thus, it is clear that even when continuous actions are used, existing approaches generally fall into grouping action into similar control inputs for agents and provide little control over explainable agent intention.
    \item The POMDP setting uses a discount factor $\gamma$ on the expected rewards.
    % When $\gamma$ is close to $1$, the rewards obtained later in the episode are weighted more heavily than when $\gamma$ is close to $0$.
    The consequence is that the rewards obtained towards the end will always be weighted less than the rewards obtained in the beginning. In adversarial settings that stretch over longer episodes, for example, head-to-head racing, chess, and Go, most of the reward is not obtained until the very end of the game. The POMDP formulation is myopic and becomes less suitable for such problems.
    \item When considering an adversarial setting in which the outcome of the competition is heavily influenced by the interaction between two agents, existing approaches, for example, Sinha et al.~\cite{sinha_formulazero_2020}, maintain a belief vector over parameterization of the opponent agent. However, the performance of such approaches does not generalize when the opponent's strategy goes out of distribution.
\end{enumerate}

% \todo{transition to solution}
To address these issues, we propose an alternative framework, specifically an alternative space, dubbed the \textbf{\textit{Objective Space}}, in which policies operate and are optimized. Instead of discretizing by grouping, we propose discretizing by \textit{abstraction} (Figure \ref{fig:method_overview}). i.e., representing policies by a combination of their properties, or their desired \textit{objectives}. Back to our racing example, a racing policy is represented by how aggressive and conservative it is instead of its parameterization. Within this framework, policies generate interpretable actions by changing the characteristics of agent policies. Furthermore, instead of optimizing for aggregated discounted rewards, we aim to minimize regret where the long-term effect of actions is considered.
In short, the overall goal is to formulate the agent's action selection problem into the following equation.
% \todo{new objective here, needs notation for new action/action space, new policy/param, new reward}
\begin{equation}\label{eq:new_obj}
\begin{split}
    \{a_0', a_1', \ldots, a_T'\}^* &= \underset{a_{t}'\sim\pi'(\theta)}{\argmin}~~\sum_{t} R_{T}(a_{t}') \\
    \pi'(\theta)&\sim \mathcal{P}_{\mathcal{A'}} \\
\end{split}
\end{equation}
Where $R_T$ is the regret generated by the sequence of actions against another agent, $\mathcal{P}_{\mathcal{A}'}$ is an empirical distribution over set $\mathcal{A}'$, a countable finite set of actions.

\subsection{Contributions}
In summary, existing frameworks do not adequately address the challenges associated with continuous space POMDPs.
The primary contribution of this work is:
\begin{enumerate}[label=(\roman*)]
    \item A agent action discretization method that \textbf{encodes interactive agents via abstraction} in a new action space dubbed the \textit{\textbf{Objective Space}}.
    % A dimension reduction formulation which \textbf{encodes interactive agents via abstraction} that better discretizes an agent's action space.
\end{enumerate}
which addresses challenges (1) and (3). The secondary contributions of this work are:
\begin{enumerate}[label=(\roman*)]
    \setcounter{enumi}{1}
    \item An efficient pipeline that \textbf{synthesizes a population of agents with multi-objective optimization}.
    \item A novel \textbf{game-theoretic planning strategy using counterfactual regret minimization with function approximation} to solve an extensive form game version of the original problem.
    % An algorithm that \textbf{predicts the counterfactual regret of a racing agent} against different optimized agents.
    % \item Lastly, \textbf{a novel game-theoretic planning strategy} that selects racing parameters using the prediction model.
\end{enumerate}
which address challenge (2).
\subsection{Related Work}
\subsubsection{RL in Latent Space}
Our proposed framework is closely related to methods that operate in a latent space. Typically, encoder architectures are used to create implicit models of the world, agent intentions, agent interactions, or dynamics.
Using a neural network to model the evolution of the world was proposed~\cite{schmidhuber_-line_1990} and recently revisited~\cite{ha_world_2018}. These approaches allow the agents to train themselves inside ``hallucinated dreams" generated by these models.
Similarly, Hafner et al.~\cite{hafner_dream_2020} learns the latent dynamics of complex dynamic systems and trains agents in latent imagination for traditionally difficult control tasks.
Schwarting et al.~\cite{schwarting_deep_2021} uses latent imagination in self-play to produce interesting agent behaviors in racing games.
There is also a line of work~\cite{kaiser_model-based_2020,zhang_solar_2019} that combines video prediction models with latent imagination for model-based RL tasks.
Finally, Xie et al.~\cite{xie_learning_2021} uses the latent space to represent the intention of the agent in multi-agent settings.
None of the above-mentioned approaches provides explainable latent spaces. In comparison, our proposed approach provides foundation for interpretable spaces where agent actions are still abstracted into a lower dimension.

\subsubsection{Regret Minimization with Approximation}
Value function approximation~\cite{sutton_reinforcement_2018, mnih_playing_2013} is widely used in reinforcement learning. Similarly, we approximate the converged counterfactual regret during the proposed approximated CFR.
Jin et al.~\cite{jin_regret_2018} approximates an advantage-like function as a proxy for regret in a single agent setting. Brown et al.~\cite{brown_deep_2018} approximates the behavior of CFR in a multi-agent setting. Our differs from both in that we directly approximate the counterfactual regret in a multi-agent setting.

\subsubsection{Value Decomposition}
When constructing the new agent action space, our approach is inspired by Value Decomposition Networks (VDN)~\cite{sunehag_value-decomposition_2017}. In VDN, the learned network decomposes value functions for a team of agents in MARL. Our proposed method decomposes long-term reward into surrogate objectives with multiple basis functions.

% \subsubsection{Different game formulations}
% There exist methods that do not discretize the agents' action space. Instead, they formulate the game as an optimal control problem under different game formulations (e.g. Differential Games).

%% file: method.tex
\section{Methodology}

\subsection{Overview}\label{sec:method_overview}
We first show an overall picture of our proposed method.

First, we discretize agent actions by decomposing agent optimization objectives into a function space with basis functions describing surrogate agent characteristics. For example, for an autonomous race car, how aggressive or conservative the driving policies are.
Then we generate the inverse of the function space by generating populations of agents using population-based multi-objective optimization. The inverse is used to define the new discretized agent actions where each action increases or decreases the function value of a single basis function of the function space.
Lastly, we form the original problem into a two-player, finite, extensive form game, and solve the game with counterfactual regret minimization (CFR) with function approximation. In the following sections, we will describe details of each module.

\begin{figure}
    \centering
    \includegraphics[trim={2cm 0 1cm 0},width=0.95\columnwidth]{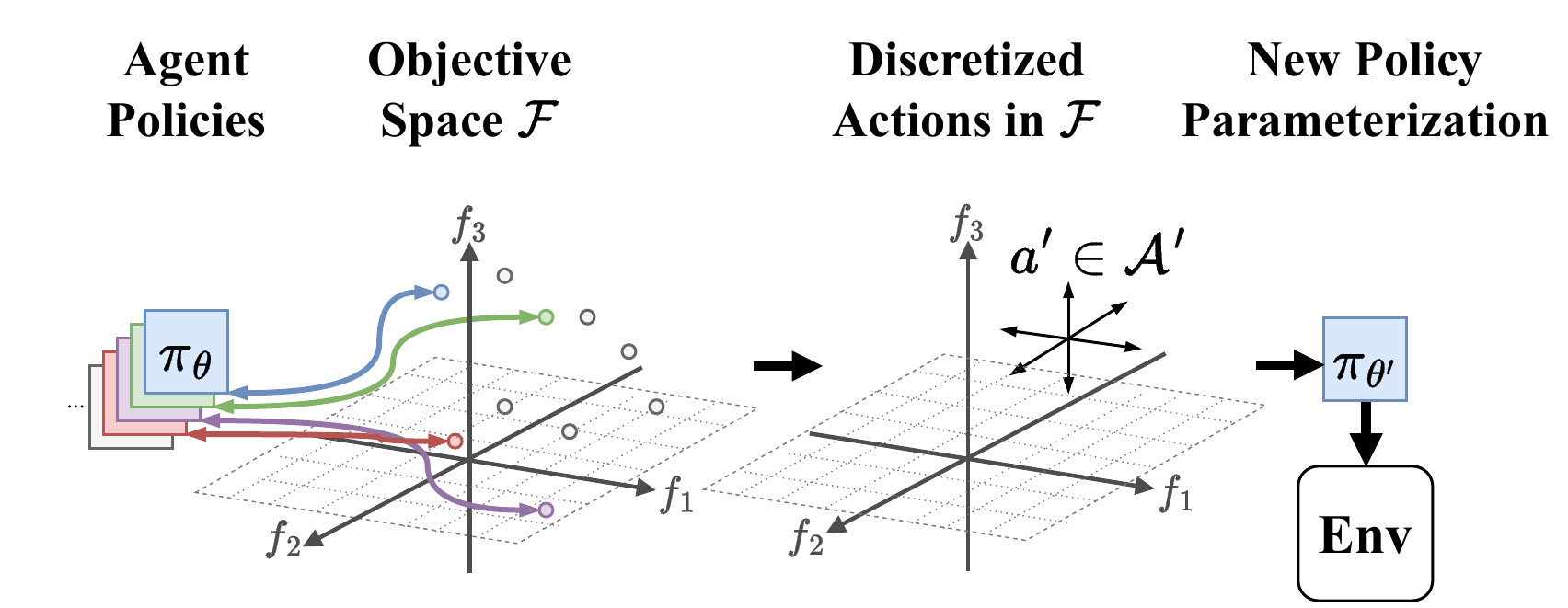}
    \caption{Agent Action Discretization. In the proposed method, agent policies are encoded into the Objective Space via designed basis functions. The new action set corresponds to moving in the Objective Space. When an action is chosen, a new policy parameterization is selected.}
    \label{fig:method_overview}
    \Description{The figure depicts the proposed agent discretization method. The image shows the proposed method where agent policies are encoded into the Objective Space via designed basis functions. The new action set corresponds to moving in the Objective Space. When an action is chosen, a new policy parameterization is selected.}
    \vspace{-10pt}
\end{figure}

\subsection{Agent Action Discretization}\label{sec:action_dis}
We take inspiration from the value decomposition~\cite{sunehag_value-decomposition_2017}. We can think of general policy optimization as ``moving'' a policy's characteristics towards being performant on a reward defined by the task. Our approach decomposes this single metric by defining multiple characterization functions $f_k:\mathcal{S\times O \times A}\rightarrow \mathbb{R}$, where each function $f_k, k\in\{1,2,\ldots,n\}$ models some property $k$ of the outcome (generated trajectory) of each agent policy. For example, in a car racing task, we can model the aggressiveness and restraint of the racing policy. We then form a function space $\mathcal{F}\subseteq\mathbb{R}^n$ using these characteristic functions as basis functions.
% In other words, we create a mapping from the policy space to $\mathcal{F}$: $[f_1,f_2,\ldots,f_n]: \mathcal{O\times A}\rightarrow \mathbb{R}^n$.
We refer to this function space as the \textbf{\textit{Objective Space}} in the following discussion.
Let the vector $V_{\mathcal{F}}=[f_1(\theta), f_2(\theta),\ldots, f_n(\theta)]$ denote a point in the Objective Space. This new function space can be related to the original policies of the POMDP as follows.
\begin{equation}\label{eq:inverse_2_policy}
    \pi(\theta) = \mathcal{F}^{-1}(V_{\mathcal{F}})
\end{equation}
For a point $V_{\mathcal{F}}$ in $\mathcal{F}$, the inverse is defined as:
\begin{equation}\label{eq:inv_def}
    \mathcal{F}^{-1}(V_{\mathcal{F}})=\underset{\theta}{\argmin}||V_{\mathcal{F}}-\mathcal{F}(\theta)||_2
\end{equation}
We describe how a population of $\theta$s are generated for the inverse in Section \ref{sec:pop_synth}. Next, we define a legal action in the new action space $\mathcal{A}'$. For a point $V_{\mathcal{F}}\in\mathbb{R}^n$ in $\mathcal{F}$, the new point induced by an action $a'(V_{\mathcal{F}}), a'\in \mathcal{A}'$ will have the following properties.
\begin{equation}\label{eq:legal_act}
\begin{split}
    \exists k\in\{1,2,\ldots,n\},~~~~~&\left| V_{\mathcal{F}}[k] - a'(V_{\mathcal{F}})[k] \right| = \epsilon, \epsilon > 0 \\
    \forall j\neq k,~~~~~&~V_{\mathcal{F}}[j] = a'(V_{\mathcal{F}})[j]
\end{split}
\end{equation}
This means that in the new action space $\mathcal{A}'$, actions are increasing or decreasing the value of only one of basis functions. Thus, such a transformation reduces the motion planning space from infinitely large to $2^n$, and also provides an explanation for agent actions, depending on how these characterization functions are defined. This addresses challenge one. Furthermore, the input of the characterization functions are state space trajectories, or production of agent policies in the state space. In a POMDP setting, state space trajectories of an adversarial agent are usually assumed to be observable. Thus, modeling opponent agent actions in this discretization setting is fairly straightforward. An overview of the method can be found in Figure \ref{fig:method_overview}. This addresses challenge two.

\subsection{Agent Population Synthesis}\label{sec:pop_synth}
\begin{figure}[h]
    \centering
    \includegraphics[trim={2cm 0 0 0},width=1\columnwidth]{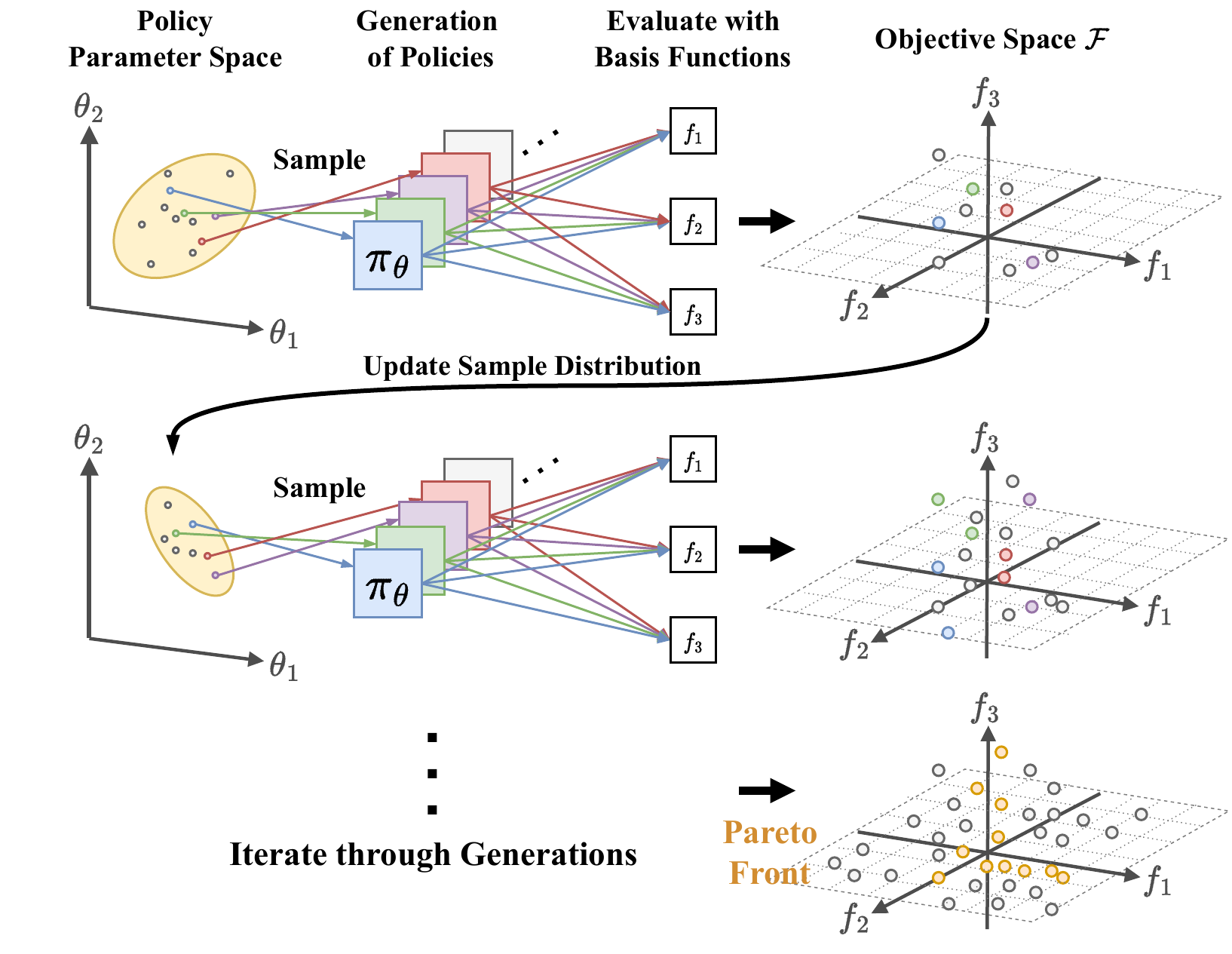}
    \caption{Population Synthesis. Population-based optimization is an iterative process that maintains a sampling distribution. At each iteration, a generation of genomes is sampled using the distribution, then put through the basis functions into the Objective Space. Based on the performance of each genome, the sampling distribution is updated. The figure depicts 2D parameter space and 3D Objective Space, but in practice there are no limits in their dimensions.}
    \label{fig:pop_synth}
    \vspace{-10pt}
    \Description{Population Synthesis. Population-based optimization is an iterative process that maintains a sampling distribution. At each iteration, a generation of genomes is sampled using the distribution, then put through the basis functions into the Objective Space. Based on the performance of each genome, the sampling distribution is updated. The figure depicts the parameter space in 2D and the Objective Space in 3D, but in practice there are no limits in their dimension.}
    \vspace{-10pt}
\end{figure}

In order to define the inverse, we need to create a discrete population of $\theta$s in Equation \ref{eq:inv_def}. Denote the set of all possible policy parameterizations as $\Theta$. Since it might not be defined at every point in the function space $\mathcal{F}$, we define the inverse loosely by generating a population of $\theta$s using population-based multi-objective optimization where the objectives are the previously defined characterization functions:
\begin{equation}\label{eq:moop_obj}
    \min_{\theta\in\Theta}(f_1(\theta),f_2(\theta),\ldots,f_n(\theta))
\end{equation}
Note that here the functions could be negated in the minimization depending on the specific characteristic function.
Since a $\theta$ that minimizes all objectives simultaneously usually does not exist, we use the Pareto Front $\Theta^*$ as the population.
\begin{equation}\label{eq:pareto}
\begin{split}
    \theta^*\in\Theta^* \iff &\forall i\in\{1,\ldots,n\}, \forall j\in\{1,\ldots,|\Theta|\}~~~f_i(\theta^*)\leq f_i(\theta_j) \\
    \text{And}~~~& \exists i\in\{1,\ldots,n\}, f_i(\theta^*)<f_i(\theta_j)
\end{split}
\end{equation}
The population synthesis process is depicted in Figure \ref{fig:pop_synth}. Since multi-objective optimization is a well studied research area and not a core contribution of this paper, we refer the readers to the literature~\cite{marler_survey_2004,hansen2016cma} for more technical discussions on the subject.

\subsection{Game Model and Notation}\label{sec:game_model}
% \todo{might need a figure for this section}
After the discretization of the actions, one could choose their policy optimization method based on the specific application. In this paper, we specifically study a two-player, zero-sum, finite game. Thus, we model it as an Extensive Form Game, which has a tree-like structure. We also make a connection to the original POMDP by also discretizing it in time. We define each transition between nodes on the game tree as a fixed-length sub-episode in the original POMDP.
% Each node in the game tree denotes a state in time in the original POMDP, including the state of the agents and the past states of the agents.
We partition the original POMDP episode $T_{0,\ldots,T}$ into $m$ segments $[T_{0,\ldots,T/m}, T_{T/m,\ldots,2T/m}, \ldots, T_{(m-1)T/m,\ldots,T}]$. At each transition point between segments, an agent is allowed to perform an action defined in Equation \ref{eq:legal_act}.
This means that the game tree will have depth $m$. An important observation here is that even with this discretization, since the state space of the original POMDP is continuous, the new game's state space is still infinite, i.e. there are infinitely many types of nodes in the game tree. We will discuss how to address this in Section \ref{sec:cfr}.

Formally, we define the simplified game as a \textit{zero sum extensive form game} with \textit{partial information} and \textit{perfect recall}. The agents have partial information since the original formulation is a POMDP, perfect recall since the agents have unlimited memory.
We denote a player as $i\in P$. History $h\in H$ is all current state information, including private information known to subsets of players, and the history of actions taken. The empty set and all prefixes of $h$ are also in $H$.
$Z\subseteq H$ denotes the set of terminal histories.
Actions are denoted as $\mathcal{A}'(h)=a'$ where $h$ is nonterminal. Note that this is the same set of actions as defined in Equation \ref{eq:legal_act}.
The information set, or infoset, $I_i\in \mathcal{I}_i$ for the player $i$ is similar to the history, but only contains information visible to the player $i$.
% \todo{might not need this next one}
% $h\in I$ is the set of all histories that appear the same in the eye of the player.\todo{end}
The strategy of the player $i$ is denoted as $\sigma_i\in\Sigma_i$, which is a distribution over actions $\mathcal{A}'(I_i)$, and $\Sigma_i$ is the set of all strategies for the player $i$.
Furthermore, the strategy is the probability of taking action $a'$ given the information set $I$.
$\sigma$ is the strategy profile that comprises all the strategies of the players. $\sigma_{-i}$ is strategies of all players except $i$.
The probability of reaching history $h$ with the strategy profile $\sigma$ is denoted by $p^{\sigma}(h)$. And $p^{\sigma}(h)_{-i}$ denotes the probability of reaching $h$ without the contribution of the player $i$.
The terminal utility, or payoff, of a player $i$ for a terminal history $h\in Z$ is $u_i(h)$.

% History includes the opponent's record of actions, while an infoset does not. In our case, a history $h$ includes the players' trajectories of positions in the objective space and the players' record of action taken at each step. The action space of each agent is increasing or decreasing a single objective value for a fixed amount $d_{\text{move}}$ at a time. In all our experiments, we use $d_{\text{move}}=1.0$. We define the terminal utility of the game as the lead $S_{\text{lead}}$ of the winning agent on the curvilinear distance along the global race line. To keep the game zero-sum, the winning agent gets utility $S_{\text{lead}}$ while the losing  agent gets $-S_{\text{lead}}$.

\subsection{Counterfactual Regret Minimization}\label{sec:cfr}
Following the definition of the game model in the previous section, it is natural to optimize the strategy of our agents in a regret-minimizing framework.
We choose Counterfactual Regret Minimization (CFR). CFR is an iterative algorithm that has a convergence bound of $O\left(\frac{1}{\sqrt{T}}\right)$. CFR minimizes overall regret by minimizing counterfactual regret, and therefore can compute a Nash equilibrium in self-play~\cite{zinkevich_regret_2007}.
However, we still have not addressed the issue of infinite game states. Inspired by value function approximation in RL approaches~\cite{sutton_reinforcement_2018,mnih_playing_2013}, we address this by approximating the counterfactual regret.
After approximating counterfactual regret, we use regret matching (RM)~\cite{hart_simple_2000} as the strategy of an iteration since it does not require parameters. We next describe the procedure of CFR with approximate counterfactual regret.

The counterfactual value $\mathbf{v}_i^\sigma(I)$ of the player $i$ is the expected utility of the player $i$ reaching $I$ with probability one.
\begin{equation}
    \label{eq:cf_val}
    \mathbf{v}_{i}^\sigma(I)=\sum_{z \in Z_{I}} p^{\sigma}_{-i}(z[I]) p^{\sigma}(z[I]\rightarrow z) u_{i}(z)
\end{equation}
Where $Z_I$ is the set of terminal histories reachable from $I$ and $z[I]$ is the prefix of $z$ up to $I$. $p^{\sigma}(z[I]\rightarrow z)$ is the probability of reaching $z$ from $z[I]$. And $\mathbf{v}_{i}^\sigma(I, a')$ follows the same calculation and assumes that player $i$ takes action $a'$ on the information set $I$ with probability one.
The immediate or instantaneous counterfactual regret is
\begin{equation}
    \label{eq:imm_cf}
    r^t(I, a') = \mathbf{v}_{i}^{\sigma^t}(I, a') - \mathbf{v}_{i}^{\sigma^t}(I)
\end{equation}
The \textit{counterfactual regret} for information set $I$ and action $a$ at iteration $t$ is
\begin{equation}
    \label{eq:cfr}
    R^t(I, a')=\sum_{\tau=1}^tr^{\tau}(I, a')
\end{equation}
Up to this point, we've condensed the calculation of counterfactual regret into a very compact form. Here, we introduce the function approximator $g_{\phi}$ parameterized by $\phi$, where we approximate the counter factual regret at iteration $t$ as:
\begin{equation}\label{eq:approx_cfr}
    R^t(I, a') \approx g_{\phi}(I, a')
\end{equation}
Additionally, we clip the counterfactual regret by using $R_{+}^t(I, a')=\max\{R^t(I, a'), 0\}$.
Regret Matching is used to pick the next action. In RM, the strategy for iteration $t+1$ is:
\begin{equation}
    \label{eq:imm_cf_strat}
        \sigma^{t+1}(I, a')=
        \frac{R_{+}^{t}(I, a')}{\sum_{a' \in \mathcal{A}(I)} R_{+}^{t}(I, a')}
\end{equation}
If the sum of the counterfactual regret of all actions at an iteration is zero, then any arbitrary strategy may be chosen.
Finally, to better cope with approximation errors~\cite{brown_deep_2018}, we choose the action with the highest approximate counterfactual regret with probability one:
\begin{equation}
    \label{eq:cf_strat}
    a'_{t+1} = \underset{a'_{t+1}\in\mathcal{A}'}{\argmax}~~~ R_+^{t+1}(I, a'_{t+1}) = \underset{a'_{t+1}\in\mathcal{A}'}{\argmax}~~~ g_{\phi}(I, a'_{t+1})
\end{equation}
Since we have the possibility to query the game using many possible combinations of action for both agents, we exploit the convergent behavior of CFR in self play.
Therefore, we set the approximation target for $g_{\phi}$ not to be the iterative behavior of CFR, but the ``converged'' counterfactual regrets. In a traditional CFR setup, the set of possible infosets is countable and finite, but the tree depth might be immense; hence, the iterative structure. In contrast, since we have partitioned the POMDP episode into equal length segments, our tree depth can be small by choice. However, our set of possible infosets is uncountable and infinite. Thus, instead of approximating the iterative behavior of CFR, we collect limited samples of full games, then directly predict the ``converged'' counterfactual regret. The next section will describe the self-play structure and how training samples are collected.

\subsection{Collecting Game Samples}
\label{sec:training}
The objective of the function approximator $g_{\phi}$ is to approximate the counterfactual regret at the final iteration of CFR. Therefore, the input of the approximator is the set of information and the action being evaluated. We set up the two player games as follows.
First, we set the depth of the game tree to a fixed integer $m$. Thus, an agent will take $m$ actions in total for the rollout. The initial starting points for both agents are chosen as random points of $V_{\mathcal{F}}$ from the Pareto Front $\Theta^*$. Then, we traverse every single branch on the game tree by taking all combinations of action at each node for both agents. At the terminal nodes on the tree, the final utilities are calculated for each game, and the corresponding counterfactual regret is also calculated for every action at every node.
% We collect the tuple $\left(\left\{a'_{\text{ego}}\right\}, \left\{a'_{\text{opp}}\right\}, R_{+}^{t}(I, a'), \right)$ for each node that is not terminal.
Since the number of actions is determined by the number of basis functions $f_k$ chosen, we know $\left|\mathcal{A}'\right|=2^k$.
With a tree depth of $m$, the total number of branches in each game tree is $2^{k(m-1)}$.
% With a tree depth of $m$, the total number of nodes in each game tree is $\frac{2^{km}}{2^k-1}$.
% However, the last levels of the game trees are terminal, no further action is taken, and counterfactual regrets are not calculated at those nodes.
The total number of games played between two agents using all possible combinations of actions will then be $2^{2k(m-1)}$.
If we select $N_{\text{init}}$ initial starting points $V_{\mathcal{F}}$ for both agents, there will be $N_{\text{init}}^2$ number of trees with different root nodes for each agent.
Thus, in total, the number of data points collected to train the approximator is $N_{\text{init}}^2 2^{2k(m-1)}$.
An information set $I$ at a specific node includes: the history of nodes traversed and the actions taken before arriving at this node, the Objective Space values $V_{\mathcal{F}}^{\text{ego}}$, $V_{\mathcal{F}}^{\text{opp}}$ of each agent of the corresponding nodes in the history,
More details on network architecture design and the training procedure will be provided in Appendix \ref{sec:app_pred}.

In summary, we started with a POMDP with continuous states, action, and observation spaces. Then, through abstraction, we discretized the agent's action space. And through partitioning the POMDP into segments of equal duration in time, we were able to optimize for the original POMDP objective in Equation \ref{eq:pomdp_obj} using CFR with counterfactual regret approximation in an extensive form game, regret minimizing setting, giving us the final action selection policy in Equation \ref{eq:cf_strat}.

% \begin{equation}\label{eq:efg_obj}
% \begin{split}
%     a'_{t+1} &= \underset{a'_{t+1}\in\mathcal{A}'}{\argmax}~~~ R_+^{t+1}(I, a'_{t+1}) \\
%     \underset{a'\in\mathcal{A}'}{\argmin} &\sum_{t=1}^T R_T(a'(v_{\mathcal{F}_{t-1}})) = \underset{a'\in\mathcal{A}'}{\argmin} \sum_{t=1}^T R_T(a'(F(\theta_{t-1}))\\
%     &= \underset{a'\in\mathcal{A}'}{\argmin} \sum_{t=1}^T R_T(a'([f_1(\theta_{t-1}), f_2(\theta_{t-1}), \ldots, f_n(\theta_{t-1})]))
% \end{split}
% \end{equation}

%% file: experiment.tex
\section{Experiments}
\begin{figure*}[t]
    \centering
    \includegraphics[width=2\columnwidth]{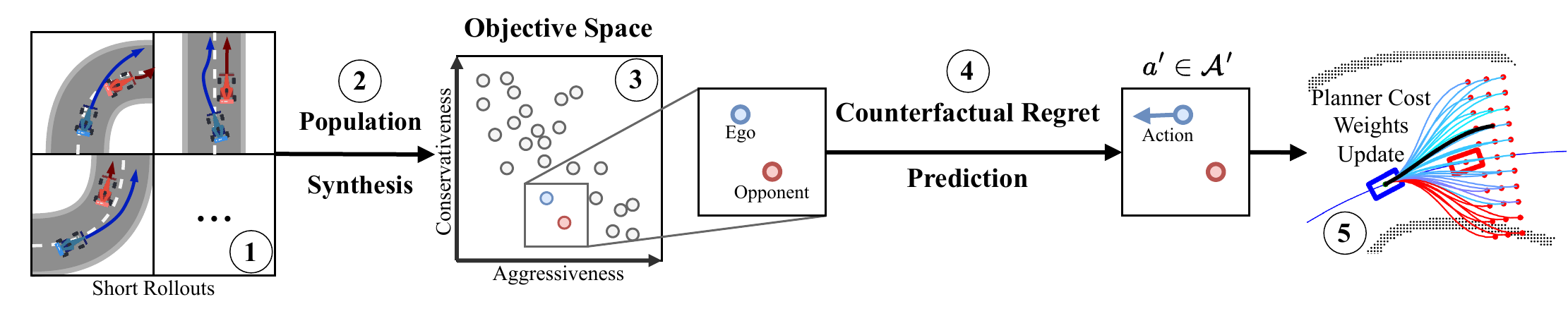}
    \caption{Overview of experiment pipeline. Simulated racing between two autonomous race cars is used as a case study (Marker 1, Section \ref{sec:sim_setup}). A competitive agent population is synthesized offline (Marker 2, Section \ref{sec:pop_synth_exp}). These agents are used to build the Objective Space using basis functions that model the aggressiveness and conservativeness of policies (Marker 3, Section \ref{sec:pop_synth_exp}). Then online, an approximate CFR where counterfactual regret is predicted with a learned model is used to find the best action in the Objective Space (Marker 4, Section \ref{sec:reg_pred_exp}). Lastly, with the updated motion planner parameterization, control inputs for the autonomous race car are produced (Marker 5, Section \ref{sec:moplan}).}
    \label{fig:exp_pipeline}
    \Description{Overview of experiment pipeline. Simulated racing between two autonomous race cars is used as a case study (Marker 1, Section \ref{sec:sim_setup}). A competitive agent population is synthesized offline (Marker 2, Section \ref{sec:pop_synth_exp}). These agents are used to build the Objective Space using basis functions that model the aggressiveness and conservativeness of policies (Marker 3, Section \ref{sec:pop_synth_exp}). Then online, an approximate CFR where counterfactual regret is predicted with a learned model is used to find the best action in the Objective Space (Marker 4, Section \ref{sec:reg_pred_exp}). Lastly, with the updated motion planner parameterization, control inputs for the autonomous race car are produced (Marker 5, Section \ref{sec:moplan}).}
\end{figure*}
\subsection{Simulation Setup}\label{sec:sim_setup}
We study a two-player head-to-head autonomous race scenario as the case study. The simulation environment~\cite{okelly_f1tenth_2020} is a gym~\cite{brockman_openai_2016} environment with a dynamic bicycle model~\cite{althoff_commonroad_2017} that considers side slip. The objective of the ego in this game is to progress further along the track than the opponent in the given amount of time without crashing into the environment or the other agent. The state space of an agent in the environment is $\textbf{x} = [x, y, \psi, s]$ where $x, y$ is the position of the agent in the world, $\psi$ is the heading angle, and $s$ is the progress indicator in the Frenet coordinate system~\cite{werling2010optimal}. The control input space of an agent is $u=[\delta, v]$ where $\delta$ is the steering angle and $v$ is the desired longitudinal velocity. The observation space of an agent is $\mathbf{r}\in\mathbb{R}^q$, where $r$ is the range measurement vector produced by a ray-marching LiDAR simulation, and $q$ is the number of laser beams.
Based on the simulation, the utility of the winner of the game is the scalar value $(s_{\text{winner}} - s_{\text{loser}})$, and the loser is the negative of that to keep the game zero-sum. If there are collisions that end the game prematurely, both agents get zero utility. Since the agents will plan its motion in a receding-horizon fashion, we need to make a clear distinction between the motion planner time steps and the POMDP time steps. Following the notation of the previous sections, the POMDP time steps will use $t$ and $T$, and the motion planning time steps will use $\tau$ and $\mathcal{T}$ in the following discussions.

\subsection{Motion Planning and Agent Parameterization}\label{sec:moplan}
\begin{figure}[b]
    \centering
    \includegraphics[width=1\columnwidth]{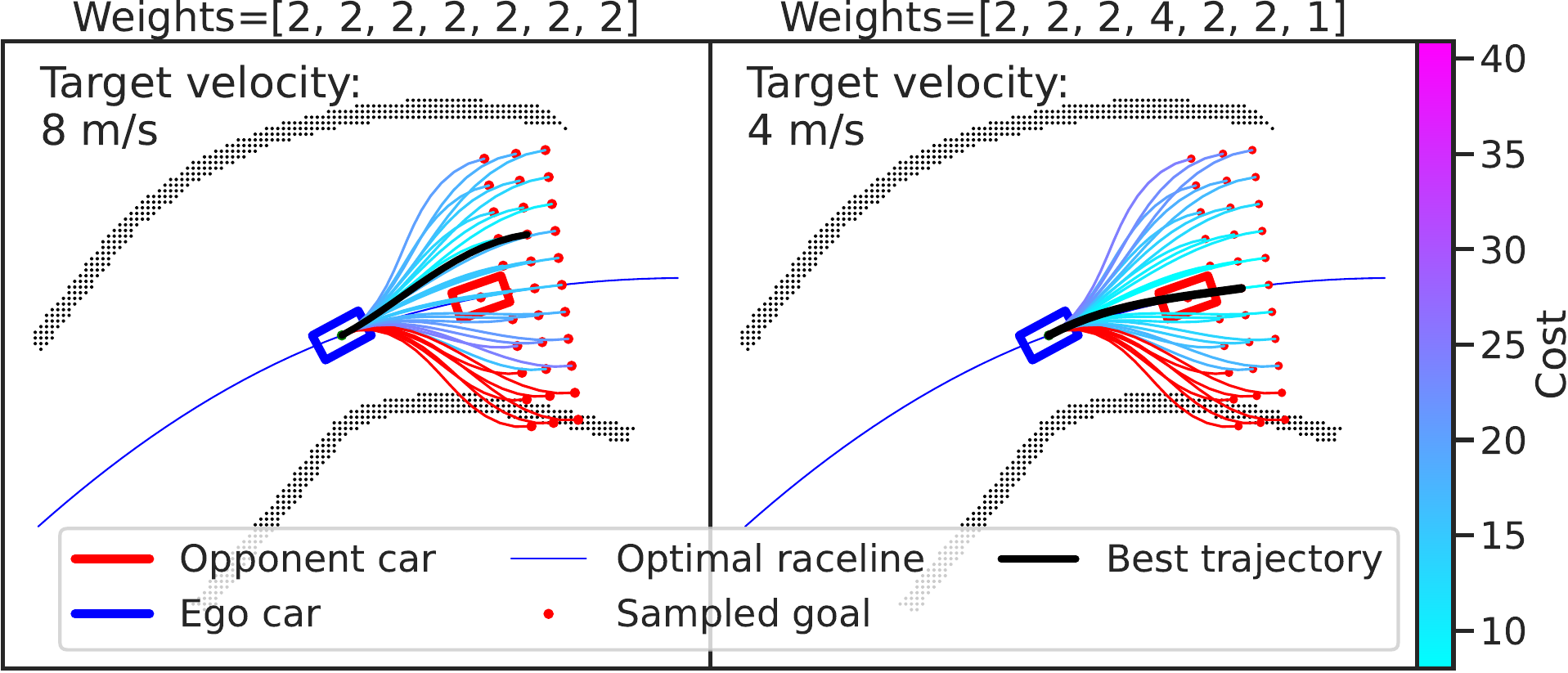}
    \caption{Effect of the different weighting of cost functions on agent behavior. A detailed description of all cost functions can be found in Appendix \ref{sec:app_planner}. The red trajectories are in collision with the track, thus assigned infinite cost.}
    \label{fig:planner}
    \Description{Figure shows a comparison between two sets of weights on cost functions when used in a lattice planner. The parameterization in the first figure weighs the penalty on deviating from a predetermined raceline less, thus generating a path that keeps a higher velocity around the opponent vehicle. The parameterization in the second figure weights this penalty more, thus generating a path that slows down and follows the opponent in front, while keeping a trajectory closely matching the raceline.}
\end{figure}
We parameterize the agent policy $\pi(\theta)$ by parameterizing the specific motion planner used by the agent. The motion planner is the link between mapping our defined agent actions in Equation \ref{eq:legal_act} to continuous control input to the agent's dynamic system.
We use a sampling-based motion planner that samples local goals for the vehicle in a lattice pattern, as shown in Figure \ref{fig:planner}. Then, a set of dynamically feasible trajectories $\Upsilon$ is generated~\cite{kelly_reactive_2003} that takes the vehicle from the current state to the sampled goal state. The predefined cost functions $c_j$ are then evaluated on all the trajectories sampled. $\theta$ in our case is a vector of weights that indicates how much the motion planner considers a certain property (maximum curvature, collision with opponent, etc.) of the proposed trajectories. The trajectory with the lowest weighted sum of the cost function is selected as the final trajectory (Equation \ref{eq:planner_cost}). More details on trajectory generation and cost function designs can be found in Appendix \ref{sec:app_planner}.
\begin{equation}\label{eq:planner_cost}
    \Upsilon^*=\underset{i\in\{1,2,\ldots,n\}}{\argmin}\sum_{j=1}^n\theta_jc_j(\Upsilon_i)
\end{equation}
After selecting a trajectory, control inputs are generated using Pure Pursuit~\cite{coulter1992implementation}. We then denote the instantaneous control input given a desired trajectory as $u_t$ defined below.
\begin{equation}\label{eq:control_in}
    u_{\tau} = \text{PP}(\Upsilon^*_\tau) = [\delta_{\tau}, v_{\tau}]
\end{equation}
For a sequence of control inputs over a horizon, we denote it as $U_{0,\mathcal{T}}=[u_0, u_1,\ldots,u_\tau]$.
As empirical evidence, we can see the different behavior of the agent induced by having different parameterizations in Figure \ref{fig:planner}. By changing how each cost function weighs in the cost calculation, the motion planner produces very different trajectories. The parameterization in the first figure weighs the penalty on deviating from a predetermined raceline less, thus generating a path that keeps a higher velocity around the opponent vehicle. The parameterization in the second figure weights this penalty more, thus generating a path that slows down and follows the opponent in front, while keeping a trajectory closely matching the raceline.

\subsection{Basis Function Definition and Population Synthesis}\label{sec:pop_synth_exp}
We next define the basis functions that form the new objective space in which we plan. In this case study, we experiment with a two-dimensional Objective Space. We define two basis functions, one modeling the aggressiveness and the other modeling the restraint of the agent.
Again, each of the basis functions takes in state space trajectories and observation histories produced by agents, and produces a scalar value. Recall from Section \ref{sec:game_model}, we have segmented the original POMDP episode into $m$ sub-episodes. To keep consistency, each basis function will be evaluated on trajectories generated during the same duration with these sub-episodes.
For aggressiveness, we design a function that measures how much an agent has made progress on the track more than its opponent. Thus, for a given trajectory, the aggressiveness is as follows.
\begin{equation}\label{eq:basis_func_agg}
\begin{split}
    f_{\text{agg}}(\theta_{\text{ego}}) = f_{\text{agg}}(\Upsilon(\theta_{\text{ego}}), \Upsilon(\theta_{\text{opp}})) = s_{\text{ego}} - s_{\text{opp}}
\end{split}
\end{equation}
For restraint, we design a function that measures how much an agent tries to avoid collision with average minimum instantaneous time-to-collision (iTTC).
And the restraint is as follows.
\begin{equation}\label{eq:basis_func_res}
\begin{split}
    f_{\text{res}}(\theta_{\text{ego}}) = f_{\text{res}}(\mathbf{r}_{\text{ego}}) = -\frac{1}{(T/m)}\sum_{t=0}^{T/m}\min_q\left[\frac{\mathbf{r}_{t,q}}{\dot{\mathbf{r}}_{t,q}}\right]_{+\infty}
\end{split}
\end{equation}
Where $\dot{\mathbf{r}}$ is the range rate and is calculated as projections of the longitudinal velocity of the vehicle to the corresponding scan angles of the LiDAR rays. The operator $[]_{+\infty}$ sets the negative elements of the vector to infinity. 
Note that both outputs of these functions depend either on the opponent agent's trajectory or on the specific track segment that the agents traveled on. We will discuss how these are selected in the population synthesis discussions. Note that for both of these basis functions, the higher the function value, the more aggressive or restraint the agent policy is. Therefore, when synthesizing for a population, we minimize the negated value of these functions (Equation \ref{eq:moop_obj}).

We synthesize the population of agents using population-based optimization. These optimization algorithms generally have the same recipe. First, a distribution of optimization variables is initialized. Then, a predetermined number (population size) of genomes are sampled from the distribution. Then each genome is evaluated for the given objectives. After collecting all the evaluation results, the sampling distribution is updated based on the results. For example, some algorithms maintain a high-dimensional Gaussian distribution, and update the mean and covariance based on the top performing samples in the last iteration. Iterating through generations, we can build a population using all previously sampled genomes. During the evaluation, we select a fixed number of random sections of the race track and a set of random parameterization $\theta$ as opponents. These random selections remain the same throughout generations. Depending on the specific use case, and desired density of the Objective Space, one can choose to subsample the population. In our case, we use a near-optimal set based on the Pareto front $\Theta^*$, denoted $\mathbb{P}_{\text{no}}$, so that the selected policies are optimized for each objective. In all experiments where a subset of initial points in the Objective Space is needed, we use a Determinantal Point Process (DPP)~\cite{kulesza_determinantal_2012} using Euclidean distances in the Objective Space for subsampling.
More details on the optimization algorithm and how we create the subsets can be found in Appendix \ref{sec:app_opt}.

\subsection{Regret Prediction Model}\label{sec:reg_pred_exp}
When collecting training samples for the approximator $g_{\phi}$, we employ the self-play structure described in Section \ref{sec:training}. First, two subsets of the near-optimal set $\mathbb{P}_{\text{DPP1}}$ and $\mathbb{P}_{\text{DPP2}}$, where $\mathbb{P}_{\text{DPP1}}\cap \mathbb{P}_{\text{DPP2}}=\varnothing$, and $\left|\mathbb{P}_{\text{DPP1}}\right| = \left|\mathbb{P}_{\text{DPP2}}\right|$. We construct game trees using each pair in the Cartesian product $\mathbb{P}_{\text{DPP1}}\times \mathbb{P}_{\text{DPP2}}$ as the initial starting points in the Objective Space of the two agents. We then play through the entire game tree and collect the necessary data points for training. We used a multilayer perceptron (MLP) with one hidden layer of size 2048 and leaky ReLU activation as $g_{\phi}$. The network is trained using L1 loss on the prediction targets, and optimized with Adam with adaptive learning rate for 2000 epochs. More details on the number of each chosen subset and the total number of data points can be found in Appendix \ref{sec:app_pred}.

To put everything together, the game-theoretic planner works in the following order. First, the ego selects a random starting point on the Pareto Front in the Objective Space. Then, the approximated CFR observes the opponent's trajectory to locate it in the Objective Space, and predicts the counterfactual regret for each available action. Next the action with highest approximate counterfactual regret is taken, and moves the ego's current position in the Objective Space to a new point. The corresponding cost weights for the motion planner are used to update the motion planner. Lastly, the motion planner generates the control input for the ego agent.

\subsection{Simulated Racing}
In the experiments, our aim is to answer the following four questions.
\begin{enumerate}
    \item Does the agent action discretization aid our agent in learning a more competitive and general policy?
    \item Does being game-theoretic improve an agent's win rate against a competitive opponent?
    \item Does the proposed agent action discretization provide interpretable explanations for agent actions?
    \item Does the proposed approach generalize to unseen environments and unseen opponents?
\end{enumerate}

\subsubsection{Action Discretization in Learning}

To answer the first question, we compare the results of a single agent environment where the objective is to finish two laps as fast as possible on the race track without crashing.
We compare three agents in this experiment.
The first is PPO~\cite{schulman_proximal_2017} with continuous actions on both steering and throttle.
The second is PPO with discretized actions: turn left, turn right, and stay straight.
The third is our proposed approach without the CFR updates.
The agents under both PPO settings are rewarded by a small keep alive reward for every time step the car is not in collision and a large terminal reward when finishing two laps under the time limit. They are also penalized by a value scaled with lap time. The PPO policies use the range measurement vector from the LiDAR scan as the observation.
During training, the PPO agent with discrete actions converges to a policy that can finish two laps. In comparison, the PPO agent with continuous actions does not.
We set up the experiments as follows. For PPO agents, the same trained agent is used in all rollouts, and the starting positions of the agents are slightly different in each rollout. For agents using the proposed approach, we again sample a subset of the near-optimal set using DPP. The number of DPP samples here matches the number of random starts for the PPO agents. We record the success rate and lap times over 20 trials for each agent both on the seen map and on the unseen map. More details on the implementation of PPO agents can be found in Appendix \ref{sec:app_ppo}.

\begin{table}[h]
\centering
\caption{Success rate and elapsed times of different agents finishing two laps in a single agent setting.}
\begin{tabular}{|c|c|c|}
\hline
Agent & Success Rate & Avg. Elapsed Time (s) \\
\hline
\multicolumn{3}{|c|}{On Seen Map} \\
\hline
PPO-continuous & 0.0 & N/A \\
PPO-discrete & 0.3 & $63.782 \pm 0.225$ \\
Ours (w/o CFR) & \textbf{1.0} & $48.533 \pm 1.398$ \\
\hline
\multicolumn{3}{|c|}{On Unseen Map} \\
\hline
PPO-continuous & 0.0 & N/A \\
PPO-discrete & 0.25 & $67.292 \pm 0.156$ \\
Ours (w/o CFR) & \textbf{1.0} & $50.814 \pm 1.343$ \\
\hline
\end{tabular}
\label{tab:rl_comp}
\end{table}
The recorded results are shown in Table \ref{tab:rl_comp}. First, we compare the two PPO agents to see the effect of using discretized actions. Our experiment confirms that agents with discretized actions are easier to train. At the end of training, the continuous PPO agent, although turning in the correct direction when encountering corners, was unable to complete the entire lap.
Then we compare the discrete PPO with our proposed method. Although able to complete the two laps during training, simply changing the starting position by a small amount drops the success rate to only 30\% even when evaluations are performed on the same map the agent saw during training. When moved to an unseen map, the success rate drops further to only 25\%.
In comparison, the proposed method maintains a perfect success rate throughout the evaluations, even when moved to an unseen map.
Although not the main objective of this experiment, we see that the proposed method was able to achieve a lower lap time across the board. Although PPO agents are awarded for having shorter lap times, they cannot compete with the proposed method.

\subsubsection{Effect of being Game-theoretic}
\begin{table}[h]
\centering
\caption{Win-rates in head-to-head racing experiments with mean win rate differences and p-values.}
\begin{tabular}{|c|c|c|c|c|}
\hline
\multirow{3}{*}{Opponent} & \multicolumn{2}{c|}{Ego} & \multirow{3}{*}{$\Delta_\mu$} & \multirow{3}{*}{p-value} \\
\cline{2-3}
 & Win-rate & Win-rate & & \\
     & Non-GT & GT & & \\
\hline 
\multicolumn{5}{|c|}{On Seen Map} \\
\hline
Non-GT & $0.515 \pm 0.251$ & $0.569 \pm 0.213$ & $0.054$ & 0.0142 \\
Random & $0.624 \pm 0.225$ & $0.670 \pm 0.199$ & $0.046$ & 0.00370 \\
Unseen & $0.586 \pm 0.101$ & $0.597 \pm 0.089$ & $0.011$ & 0.0863 \\
\hline
\multicolumn{5}{|c|}{On Unseen Map} \\
\hline
Non-GT & $0.553 \pm 0.256$ & $0.628 \pm 0.180$ & $0.075$ & 0.0124 \\
Random & $0.625 \pm 0.278$ & $0.738 \pm 0.172$ & $0.113$ & 0.00276 \\
Unseen & $0.556 \pm 0.101$ & $0.565 \pm 0.098$ & $0.009$ & 0.147 \\
\hline
\end{tabular}
\label{tab:online_exp}
\end{table}
To answer the second question, and in order to show the effectiveness of game-theoretic planning, we race the policies from our framework against various opponents in different environments. Each experiment is a single head-to-head race with a fixed duration. We designate the winner of the race as the agent who progressed further down the track at the end of the duration. To ensure fairness, the agents start side by side at the same starting line on track, and alternate starting positions. The starting line is also randomized five times for one pair of agents. There are three types of agents in the experiment. The GT agent is produced under the proposed population synthesis framework with CFR and counterfactual regret approximation. The non-GT agent is also produced in the same population synthesis framework but without CFR. The random agent is a random selection from all the explored parameterizations from the population synthesis framework without CFR. Lastly, the unseen agent is a competitive motion planner under a completely different parameterization. The ego agents in the experiments are GT and non-GT agents, while the opponents are non-GT, random, and unseen (Table \ref{tab:online_exp}). We choose 20 different variants of each agent. Thus, each table cell in Table \ref{tab:online_exp} consists of statistics from $20^2*2*5=4000$ head-to-head racing games. Each agent is allowed $m=4$ actions, with each sub-episode lasting $T/m=8$ seconds. The total length of the games is $40$ seconds, with the first $8$ seconds as the initial sub-episode to observe the opponent.
We report the results of paired t-tests against all opponents with the null hypothesis that the use of the proposed CFR process does not change the win rate.
As shown in Table \ref{tab:online_exp}, the main result of this experiment is that the p-value is small enough to reject the null hypothesis in most cases. Across all pairings of ego and opponent, there is an increase in average win rate by using our proposed approximated CFR, significantly in most cases. Although not as significant when playing an unseen opponent, the trend is still clearly present. This finding validates our counterfactual regret approximator and the effect of CFR by showing significant improvement.

\subsubsection{Agent Characterization}
\begin{figure}[h]
    \centering
    \includegraphics[trim={2cm 1cm 1cm 2cm},width=\columnwidth]{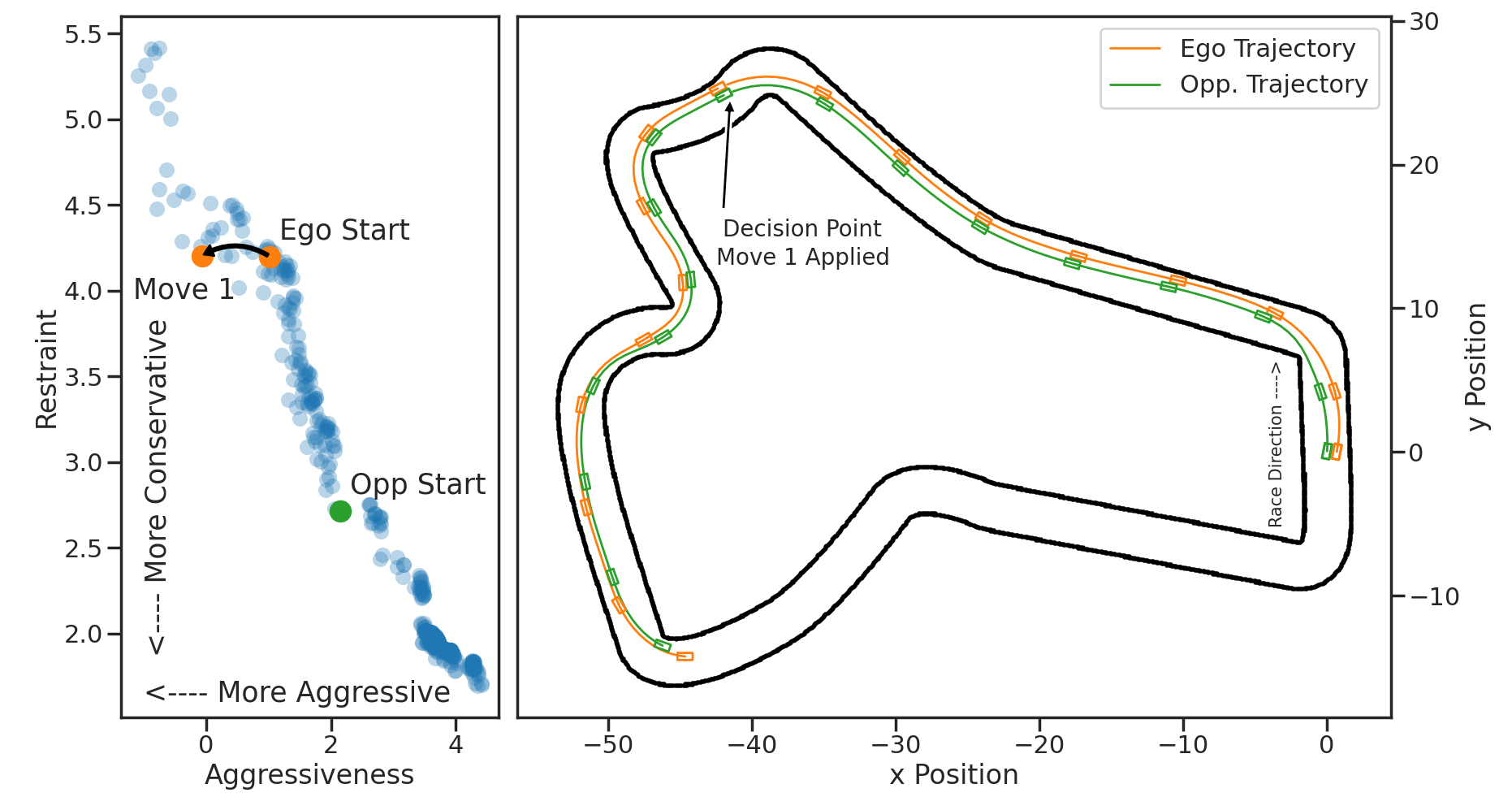}
    \caption{Effect of taking an action in the Objective Space. The left subplot shows what the action looks like in the Objective Space. The right subplot shows that after taking an action, the ego overtakes the opponent.}
    \label{fig:make_one_move}
    \Description{Figure shows the effect of taking an action in the Objective Space. The left subplot shows what the action looks like in the Objective Space. The right subplot shows that after taking an action, the ego overtakes the opponent.}
\end{figure}

% Another hypothesis of this paper is that using the proposed agent action discretization in the Objective Space provides explainable agent actions.
To answer the third question, we examine selected rollouts of the racing games to investigate whether the actions of the agents are interpretable.
In the first segment of the track (before the decision point) in Figure \ref{fig:make_one_move}, the ego agent uses a random choice of starting parameterization to plan and observe the opponent for $T/m$ seconds. The ego and the opponent remained side by side until the second corner. Then at the decision point, which is the moment we select an action and transition from one node on the game tree to the next, 
the ego observed that the opponent's strategy produced a trajectory that is positioned in the lower right quadrant of the Objective Space (more conservative than aggressive). The counterfactual regret approximator then predicted that increasing aggressiveness has the highest estimated counterfactual regret. The planner switches to the new cost weights corresponding to the new point in the objective space (left subplot). The selected action's effect is immediately evident since the ego slowed down less than the opponent in the chicane (after the decision point) and overtook the opponent by the end of the second sub-episode.

\begin{figure}[h]
    \centering
    \includegraphics[width=\columnwidth]{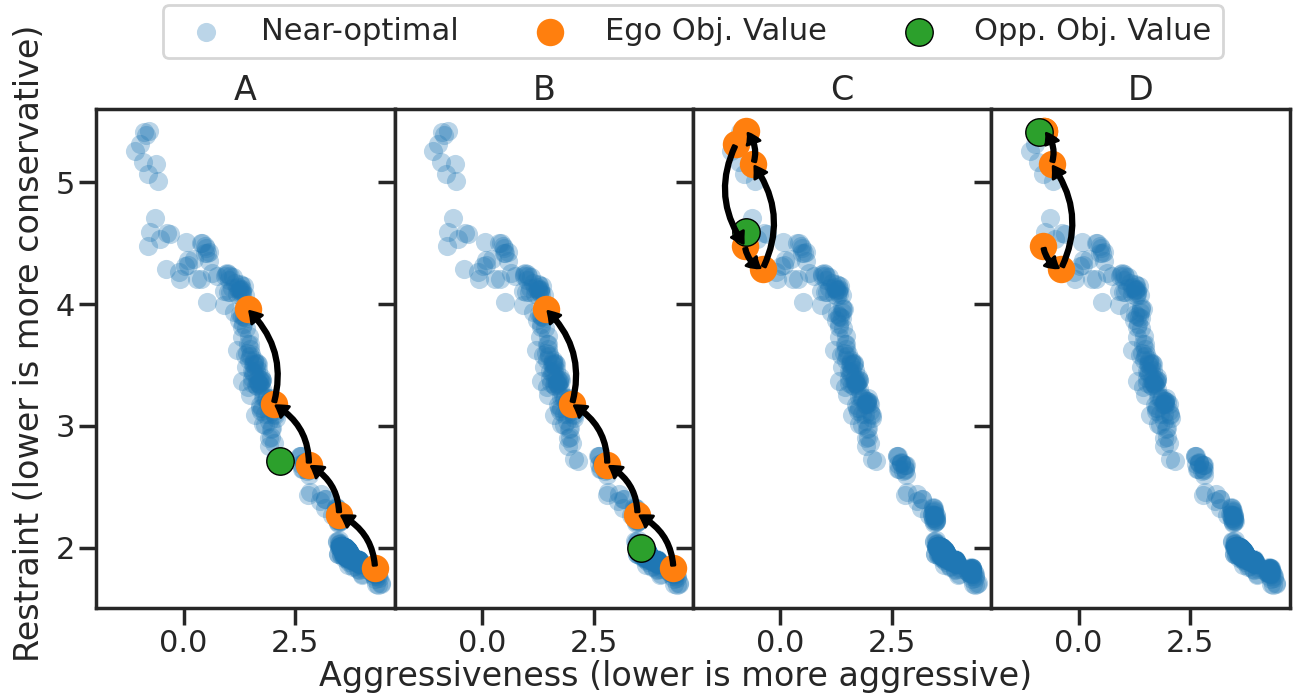}
    \caption{Trajectories of ego's actions in Objective Space. In subplots A and B, the opponent is more conservative and the ego decides to increase aggressiveness right away. In subplots C and D, the opponent is more aggressive and the ego decides to increase in restraint until there is an opportunity to increase in aggressiveness and overtake.}
    \label{fig:moves}
    \Description{Trajectories of ego's actions in Objective Space. In subplots A and B, the opponent is more conservative and the ego decides to increase aggressiveness right away. In subplots C and D, the opponent is more aggressive and the ego decides to increase in restraint until there is an opportunity to increase in aggressiveness and overtake.}
\end{figure}

We then take a step back and look at a bigger picture which shows agent decisions of the entire rollout in the Objective Space. In Figure \ref{fig:moves}, four different rollouts are shown in which the ego wins at the end.
In subfigures A and B, the opponent agent is observed to be in the lower right quadrant of the Objective Space, meaning that these agents value safety more than progress. In these scenarios, the ego agent, starting in the lower right quadrant, became more aggressive than conservative with each action.
In subfigures C and D, the opponent agent is observed to be in the upper left quadrant of the objective space, meaning that these agents value progress more than safety. In these scenarios, the ego decides to stay conservative and only increases aggressiveness later on when there is an opportunity to overtake.

From these two closer examinations of agent actions, using interpretable basis functions to construct the Objective Space has a clear benefit. Discretizing agent actions by abstraction provides a much clearer context for explaining agent decision-making.

\subsubsection{Facing Unknown Opponents}
And lastly, to answer the fourth question, we examine the results when faced with unknown environments.
Since we defined the Objective Space basis function to take the outcome (generated state space trajectories) of a policy as input, agents deal with unseen adversaries in unseen environments better.
In the adversarial multi-agent environment from rows 3 and 6 in Table \ref{tab:online_exp}, we see that the increase in the win rate is retained, though not as significant as other scenarios.

%% file: discussion.tex
\section{Limitations and Future Work}

% First, we chose 8-second segments of races as our short adversarial rollouts. This choice encouraged agent interactions and led to manageable simulation times. Online, each action in the game also consists of 8-second segments to match the episodes seen before. The planner could implement a receding horizon scheme and use the last 8 seconds instead when playing the game. This way, the game-theoretic component can react to the changes in the environment much faster.

First, we chose to partition the original POMDP episode into equal duration segments. This choice led to a distinct game tree structure with depth $m$, and kept the input space of the counterfactual regret approximator $g_\phi$ manageable. If agent decision making, or strategy switching, is desired at a higher frequency, one might employ a receding horizon update scheme. However, this might lead to too large a tree depth $m$ when designing the function approximator. Future work should focus on addressing this issue by finding the balance point for this trade-off.

\balance

Second, experiments should investigate biases in the evaluation scenarios during the optimization and experiments. Randomness affects the selection of the opponent set and the track sections during optimization. The biases in selecting evaluation scenarios are also present in the first column of Table \ref{tab:online_exp}. When a non-GT agent plays against a non-GT agent, the ego win rate should be close to 50\%. Future work can focus on designing experiments to evaluate whether this deviation from the 50\% win rate is significant. Furthermore, biases must be eliminated when selecting random subsets of the agent population and evaluating scenarios. 

Third, instead of playing against randomized opponents during population-based optimization, the opponent set should become more and more competitive as the optimization iterates. The authors experimented with mixing agents on the Pareto front periodically into the opponent set, resulting in premature convergence to less competitive agents. An alternative is to incorporate self-play into the acceptance of new genomes. Instead of accepting all new genomes in each generation, new genomes are compared against the ones from the previous generation. Only winning genomes are passed on to the next generation.

Fourth, opponents in the experiments do not adapt their planner cost weights like ego agents. Using an adaptive planner would increase competition for a fairer two-sided interaction.

Lastly, the basis functions used in the case study are chosen by hand with expert knowledge of the subject matter. On the basis of the proposed formulation, this should not be a hard requirement for improvement in agent action discretization. Discovering these basis functions automatically and selecting the metrics most different from each other to construct the Objective Space will create an interesting research problem where hidden mechanics of agent interaction could be discovered.

\section{Conclusion}
To overcome the challenges that come with continuous-space POMDPs, we propose an agent action discretization that encodes policy characteristics into the Objective Space. Agent actions produced in this space are interpretable and help with generalization.
The central hypotheses of this paper are that using the proposed discretization of agent action and CFR with counterfactual regret approximation not only significantly improves the win rate against different opponents, improves interpretability when explaining agent decisions, but also transfers to unseen opponents in an unseen environment.
We first define the Objective Space and legal actions in this space, then perform agent population synthesis via multi-objective optimization. Next, we train a counterfactual regret approximator and implement an online planning pipeline that uses CFR to maximize the win rate. Lastly, we provide statistical evidence showing significant improvements to the win rate that are generalized to unseen environments. Moreover, we provide an examination on how the agent action discretization method improves interpretability when explaining agent decisions. We significantly (with a p-value less than 0.05) improved the win rate by 5.4\% on the seen map and 7.5\% on the unseen map on average against non-game-theoretic opponents; the win rate by 4.6\% on the seen map and 11.3\% on the unseen map on average against random opponents. Lastly, we improved the win rate by 1.1\% on the seen map and 0.9\% on the unseen map against unseen opponents.

%% file: impact.tex
\section{Societal Impact}
Understanding the intent of a learning-based agent and its peers in the environment is crucial to deploying autonomous systems in adversarial multi-agent settings. Real-world applications, such as autonomous vehicles, financial decision making algorithms, and recommendation algorithms, are examples of such autonomous systems. Without a proper system that supports explaining decisions made by such systems in a human-interpretable way, it is impossible to assign blame when malfunctions occur. Especially in life-critical applications, these requirements become even more important. This work provides a preliminary framework for interpretable agent actions in autonomous systems.

%% file: appendix.tex
\appendix

\section{Motion Planner}
\label{sec:app_planner}
We use a sampling-based hierarchical motion planner similar to that of \cite{ferguson_motion_2008}. At the top level, the planner receives information on the current poses and velocities of the ego and opponent, as well as an optimal race line generated using \cite{heilmeier_minimum_2020} for the current map. The race line consists of waypoints as tuples of $(x, y, \theta, v)$, which are the desired position, heading, and velocity of the vehicle.
Then $n$ uniform grid points representing local goals are sampled around the race line (see Figure \ref{fig:planner}). 
Then dynamically feasible trajectories are generated from the current pose of the car to the sampled goals using third-order clothoids combining methods from \cite{nagy_trajectory_nodate,kelly_reactive_2003,howard_optimal_2007,mcnaughton_motion_2011,howard_adaptive_2009}.
Additionally, for each generated trajectory, $m$ velocity scaling factors are assigned to generate different velocity profiles for the same path.
Hence, the planner samples $n\times m$ trajectories at one planning step.

We define multiple cost functions ($c_j$ in Equation \ref{eq:planner_cost}) to evaluate the quality of each trajectory for geometric properties and velocity profiles. We use the following cost functions:
\begin{itemize}
    \item $c_{\text{mc}}=\max_{s=0}^{s_f}\{\kappa_s\}$: maximum curvature on the trajectory
    \item $c_{\text{al}} = $ arc length $s_f$ of the trajectory
    \item $c_{\text{hys}} = \left|\left|\mathcal{T}_t - \mathcal{T}^*_{t-1}\right|\right|_2$ hysteresis loss that measures similarity to the previously selected trajectory. Calculated as the Euclidean distance between the two trajectories. Note that the previously selected trajectory is shifted forward to compensate for the vehicle's motion between time steps.
    \item $c_{\text{do}}$ distance to the optimal race line measured by lateral deviation. For every single position on the trajectory, a corresponding nearest point is found on the optimal raceline. Then the lateral is calculated in the Frenet coordinate frame.
    \item $c_{\text{co}}$ fixed collision cost with the opponent discounted by relative speed to the opponent at each time step. 
    \item $c_{\text{v1}}$ velocity cost that encourages higher speed.
    \item $c_{\text{v2}}$ velocity cost that penalizes co-occurrence of high speed and high curvature.
\end{itemize}
In addition, we include a global velocity scaling factor $\gamma$ as another parameter for the agent. Therefore, each agent can be parameterized by the weight vector $\theta$:
\begin{equation*}
    \theta=\left[ \gamma, ~c_{mc}, ~c_{al}, ~c_{\text{hys}}, ~c_{do}, ~c_{co}, ~c_{v1}, ~c_{v2} \right]
\end{equation*}
In addition, trajectories in collision with the environment are assigned infinite costs.

\section{Population-based Agent Optimization}
\label{sec:app_opt}
We use the Covariance Matrix Adaptation Evolutionary Strategy (CMA-ES)~\cite{hansen2016cma} as the population-based multi-objective optimizer. Parameterization of each genome in the ES has the seven cost weights ranging from $1.0$ to $10.0$ and the velocity discount factor ranging from $0.6$ to $1.0$. The duration of each rollout is 8 seconds in the evaluation. In the evaluation to obtain the Objective Space position of each genome, we use 120 pairs of opponents with randomized cost weights and randomized section of the race track. The average basis function values across all pairs obtained for each genome are used to put them into the Objective Space. To encourage exploration, during every rollout, if the ego overtakes the opponent, the aggressiveness value of the genome increases by 10\%. If the ego crashes into the opponent, the aggressiveness value increases by 10\% and the conservativeness value increases by 1.

\begin{figure}
    \centering
    \includegraphics[width=1.0\columnwidth]{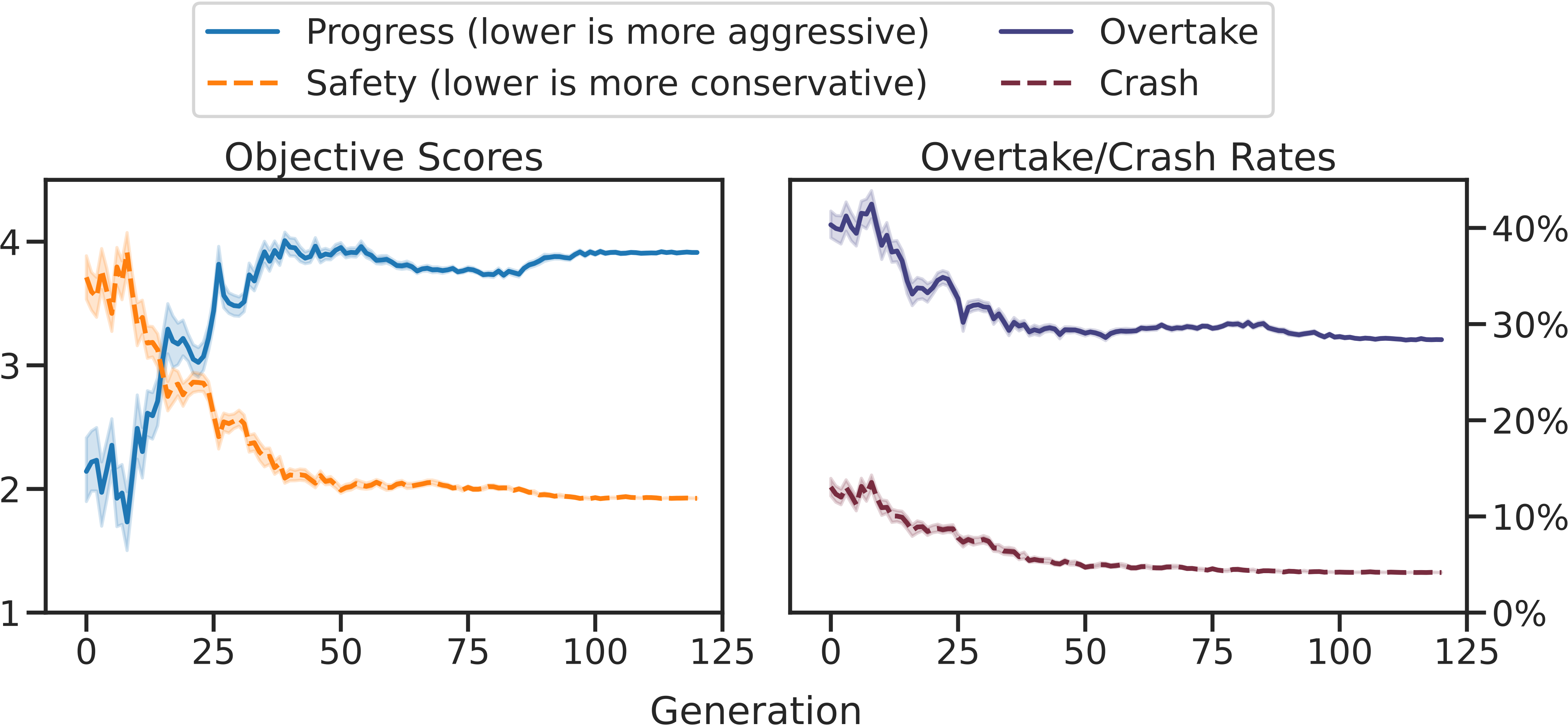}
    \caption{Progression of CMA-ES optimization. The x-axis denotes generations of 100 genomes.}
    \label{fig:offline_result}
\end{figure}

In Figure \ref{fig:offline_result}, we show the progression of the two competing objectives and the crash and overtake rate during optimization. The figure shows that the overtake rate is higher when the current parameterization is more aggressive. However, the crash rate is also higher. The trend of the objective scores in the figure shows that the two objectives are competing, and we have explored a wide range of different parameterizations.

After obtaining the Pareto front $\mathbb{P}_{\text{pf}}$ and the set of all agents explored $\mathbb{P}_{\text{all}}$ from CMA-ES, we create several different subsets of the agents explored using the Objective Space.
First, we create a near-optimal set $\mathbb{P}_{\text{no}}$ using all points in the agents explored within $d_{\text{near}}=0.3$ away in Euclidean distance from every point in $\mathbb{P}_{\text{pf}}$.
Next, we use a Determinantal Point Process (DPP) to sample $N_{\text{DPP}i}=20$ samples from the near-optimal set into $\mathbb{P}_{\text{DPP}i}$. Multiple DPP subsets are used in our experiments.

\begin{figure}
    \centering
    \includegraphics[width=1.0\columnwidth]{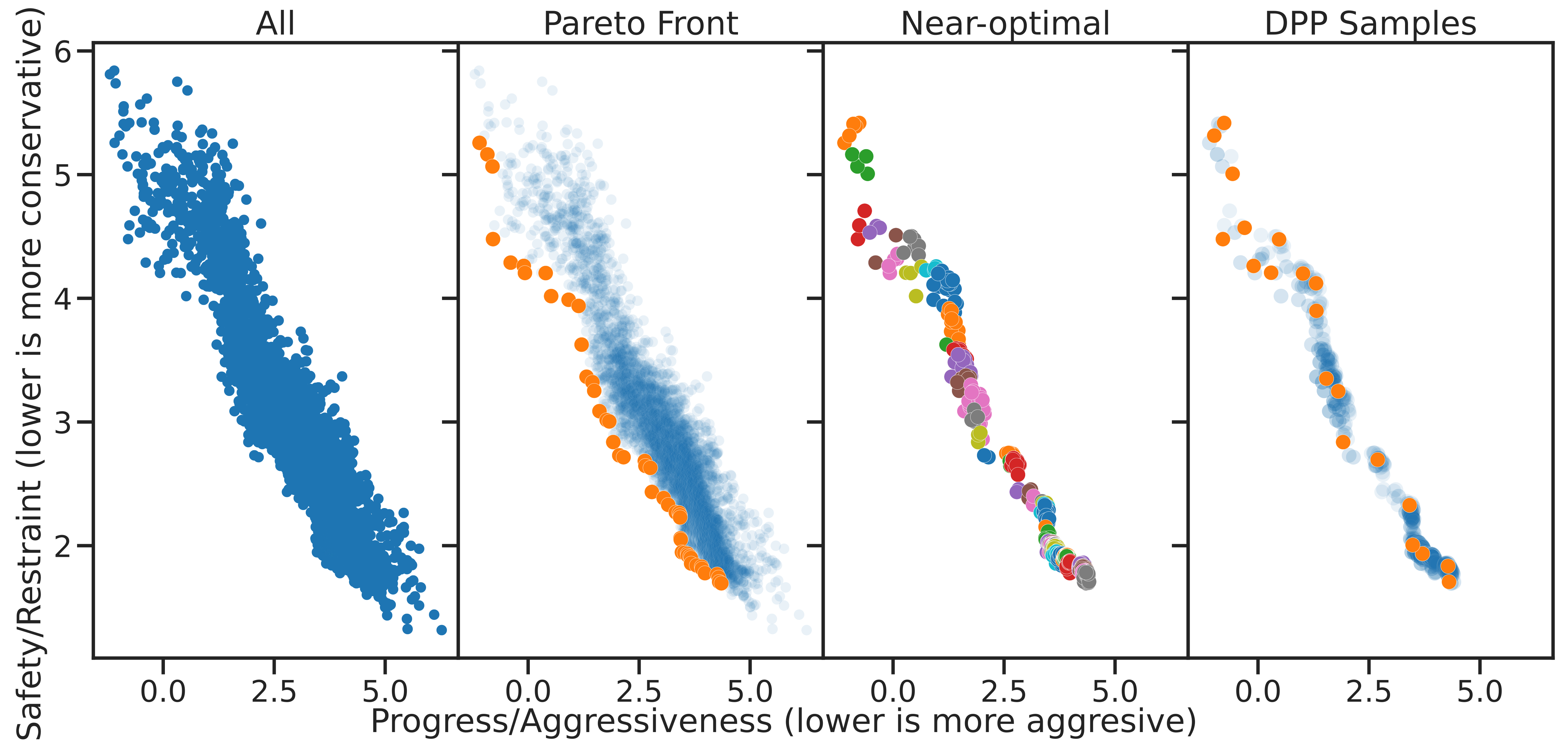}
    \caption{Position of different sets of prototypes in the Objective Space.}
    \label{fig:pareto}
\end{figure}

Figure \ref{fig:pareto} shows the subsets of prototypes $\mathbb{P}_{\text{all}}$, $\mathbb{P}_{\text{pf}}$, $\mathbb{P}_{\text{no}}$, and an example DPP subset $\mathbb{P}_{\text{DPP1}}$ in the 2D Objective Space. This figure shows the coverage of explored agents in the Objective Space.

\begin{figure}
    \centering
    \includegraphics[width=1.0\columnwidth]{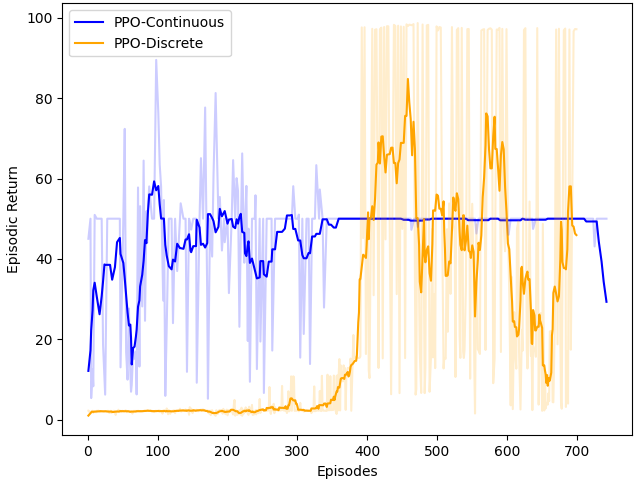}
    \caption{Episodic returns when training PPO policies. Moving average of 10 steps shown.}
    \label{fig:ppo_return}
\end{figure}

\section{Counterfactual Regret Prediction}
\label{sec:app_pred}
In our experiments, we choose our game tree depth as $m=4$ and the dimension of the Objective Space as $k=2$. The total number of branches is $4^3=64$. The possible combinations of actions taken between two agents are $64^2=4096$ games. When $N_{\text{init}}=20$, the total number of games played in data collection is $20^2*4096=1638400$. We calculated and collected the counterfactual regret at each data point for querying. When two nodes are on the same branch of the game tree, they have the same terminal utility.

 We encode Objective Space values into fixed-length vectors with zero padding. Masks are set up so that only positions with objective values are filled with ones; otherwise, zeros. All actions are one-hot encoded. In total, the input feature size for the prediction model is 40. We use a multilayer perceptron (MLP) with one hidden layer of size 2048 and leaky ReLU activation between layers. We train the network with L1 loss, batch size of 1024, and Adam optimizer with adaptive learning rate (starting at 0.005, reduce when validation loss plateaus for ten steps) in 2000 epochs.

\section{PPO Agents}
\label{sec:app_ppo}
RL agents used for the comparison are trained using continuous and discrete PPO in CleanRL~\cite{huang2022cleanrl}. We show the episodic returns for each agent in \ref{fig:ppo_return}. The training is capped at 1000000 total time steps, with 7000 steps between each policy update.
The learning rate is annealed and starts at 3e-4 for continuous PPO and 2.5e-4 for discrete PPO.
The discount factor for both is set at 0.99.
The $\lambda$ for the general advantage estimation for both is set at 0.95.
The epochs to update is 10 for the continuous policy and 4 for the discrete policy.
The surrogate clipping coefficient is 0.2 for both policies.
The entropy coefficient is 0 for the continuous policy and 0.01 for the discrete policy.
The coefficient for the value function is 0.5 for both policies.
The maximum norm for gradient clipping is 0.5 for both policies.
The continuous action is sampled from a learned Gaussian with control input limits. The discrete action is steering with 0.3, -0.3, and 0.0 steering angles, all with a velocity of 5.0 m/s.
Agents get a reward of 0.01 for every time step not in collision, a reward of 40 for reaching the end of two laps without collision. The gym environment is set with a timeout to avoid infinite episodes.

%% file: main.bbl

\begin{thebibliography}{36}

%%% the Web address.

\ifx \showCODEN    \undefined \def \showCODEN     #1{\unskip}     \fi
\ifx \showDOI      \undefined \def \showDOI       #1{#1}\fi
\ifx \showISBNx    \undefined \def \showISBNx     #1{\unskip}     \fi
\ifx \showISBNxiii \undefined \def \showISBNxiii  #1{\unskip}     \fi
\ifx \showISSN     \undefined \def \showISSN      #1{\unskip}     \fi
\ifx \showLCCN     \undefined \def \showLCCN      #1{\unskip}     \fi
\ifx \shownote     \undefined \def \shownote      #1{#1}          \fi
\ifx \showarticletitle \undefined \def \showarticletitle #1{#1}   \fi
\ifx \showURL      \undefined \def \showURL       {\relax}        \fi
% invisible to TeX
\providecommand\bibfield[2]{#2}
\providecommand\bibinfo[2]{#2}
\providecommand\natexlab[1]{#1}
\providecommand\showeprint[2][]{arXiv:#2}

\bibitem[\protect\citeauthoryear{Althoff, Koschi, and Manzinger}{Althoff et~al\mbox{.}}{2017}]%
        {althoff_commonroad_2017}
\bibfield{author}{\bibinfo{person}{Matthias Althoff}, \bibinfo{person}{Markus Koschi}, {and} \bibinfo{person}{Stefanie Manzinger}.} \bibinfo{year}{2017}\natexlab{}.
\newblock \showarticletitle{{CommonRoad}: {Composable} benchmarks for motion planning on roads}. In \bibinfo{booktitle}{\emph{2017 {IEEE} {Intelligent} {Vehicles} {Symposium} ({IV})}}. \bibinfo{pages}{719--726}.
\newblock
\urldef\tempurl%
\url{https://doi.org/10.1109/IVS.2017.7995802}
\showDOI{\tempurl}


\bibitem[\protect\citeauthoryear{Brockman, Cheung, Pettersson, Schneider, Schulman, Tang, and Zaremba}{Brockman et~al\mbox{.}}{2016}]%
        {brockman_openai_2016}
\bibfield{author}{\bibinfo{person}{Greg Brockman}, \bibinfo{person}{Vicki Cheung}, \bibinfo{person}{Ludwig Pettersson}, \bibinfo{person}{Jonas Schneider}, \bibinfo{person}{John Schulman}, \bibinfo{person}{Jie Tang}, {and} \bibinfo{person}{Wojciech Zaremba}.} \bibinfo{year}{2016}\natexlab{}.
\newblock \bibinfo{title}{{OpenAI} {Gym}}.
\newblock
\newblock
\urldef\tempurl%
\url{https://doi.org/10.48550/arXiv.1606.01540}
\showDOI{\tempurl}
\newblock
\shownote{arXiv:1606.01540 [cs].}


\bibitem[\protect\citeauthoryear{Brown, Lerer, Gross, and Sandholm}{Brown et~al\mbox{.}}{2018}]%
        {brown_deep_2018}
\bibfield{author}{\bibinfo{person}{Noam Brown}, \bibinfo{person}{Adam Lerer}, \bibinfo{person}{Sam Gross}, {and} \bibinfo{person}{Tuomas Sandholm}.} \bibinfo{year}{2018}\natexlab{}.
\newblock \showarticletitle{Deep {Counterfactual} {Regret} {Minimization}}.
\newblock  (\bibinfo{date}{Nov.} \bibinfo{year}{2018}).
\newblock
\urldef\tempurl%
\url{https://doi.org/10.48550/arXiv.1811.00164}
\showDOI{\tempurl}


\bibitem[\protect\citeauthoryear{Coulter et~al\mbox{.}}{Coulter et~al\mbox{.}}{1992}]%
        {coulter1992implementation}
\bibfield{author}{\bibinfo{person}{R~Craig Coulter} {et~al\mbox{.}}} \bibinfo{year}{1992}\natexlab{}.
\newblock \bibinfo{booktitle}{\emph{Implementation of the pure pursuit path tracking algorithm}}.
\newblock \bibinfo{publisher}{Carnegie Mellon University, The Robotics Institute}.
\newblock


\bibitem[\protect\citeauthoryear{Ferguson, Howard, and Likhachev}{Ferguson et~al\mbox{.}}{2008}]%
        {ferguson_motion_2008}
\bibfield{author}{\bibinfo{person}{Dave Ferguson}, \bibinfo{person}{Thomas~M. Howard}, {and} \bibinfo{person}{Maxim Likhachev}.} \bibinfo{year}{2008}\natexlab{}.
\newblock \showarticletitle{Motion planning in urban environments}.
\newblock \bibinfo{journal}{\emph{Journal of Field Robotics}} \bibinfo{volume}{25}, \bibinfo{number}{11-12} (\bibinfo{year}{2008}), \bibinfo{pages}{939--960}.
\newblock
\showISSN{1556-4967}
\urldef\tempurl%
\url{https://doi.org/10.1002/rob.20265}
\showDOI{\tempurl}
\newblock
\shownote{\_eprint: https://onlinelibrary.wiley.com/doi/pdf/10.1002/rob.20265.}


\bibitem[\protect\citeauthoryear{Ha and Schmidhuber}{Ha and Schmidhuber}{2018}]%
        {ha_world_2018}
\bibfield{author}{\bibinfo{person}{David Ha} {and} \bibinfo{person}{Jürgen Schmidhuber}.} \bibinfo{year}{2018}\natexlab{}.
\newblock \showarticletitle{World {Models}}.
\newblock  (\bibinfo{date}{March} \bibinfo{year}{2018}).
\newblock
\urldef\tempurl%
\url{https://doi.org/10.5281/zenodo.1207631}
\showDOI{\tempurl}
\newblock
\shownote{arXiv:1803.10122 [cs, stat].}


\bibitem[\protect\citeauthoryear{Hafner, Lillicrap, Ba, and Norouzi}{Hafner et~al\mbox{.}}{2020}]%
        {hafner_dream_2020}
\bibfield{author}{\bibinfo{person}{Danijar Hafner}, \bibinfo{person}{Timothy Lillicrap}, \bibinfo{person}{Jimmy Ba}, {and} \bibinfo{person}{Mohammad Norouzi}.} \bibinfo{year}{2020}\natexlab{}.
\newblock \bibinfo{title}{Dream to {Control}: {Learning} {Behaviors} by {Latent} {Imagination}}.
\newblock
\newblock
\urldef\tempurl%
\url{https://doi.org/10.48550/arXiv.1912.01603}
\showDOI{\tempurl}
\newblock
\shownote{arXiv:1912.01603 [cs].}


\bibitem[\protect\citeauthoryear{Hansen}{Hansen}{2016}]%
        {hansen2016cma}
\bibfield{author}{\bibinfo{person}{Nikolaus Hansen}.} \bibinfo{year}{2016}\natexlab{}.
\newblock \showarticletitle{The CMA evolution strategy: A tutorial}.
\newblock \bibinfo{journal}{\emph{arXiv preprint arXiv:1604.00772}} (\bibinfo{year}{2016}).
\newblock


\bibitem[\protect\citeauthoryear{Hart and Mas-Colell}{Hart and Mas-Colell}{2000}]%
        {hart_simple_2000}
\bibfield{author}{\bibinfo{person}{Sergiu Hart} {and} \bibinfo{person}{Andreu Mas-Colell}.} \bibinfo{year}{2000}\natexlab{}.
\newblock \showarticletitle{A {Simple} {Adaptive} {Procedure} {Leading} to {Correlated} {Equilibrium}}.
\newblock \bibinfo{journal}{\emph{Econometrica}} \bibinfo{volume}{68}, \bibinfo{number}{5} (\bibinfo{year}{2000}), \bibinfo{pages}{1127--1150}.
\newblock
\showISSN{1468-0262}
\urldef\tempurl%
\url{https://doi.org/10.1111/1468-0262.00153}
\showDOI{\tempurl}
\newblock
\shownote{\_eprint: https://onlinelibrary.wiley.com/doi/pdf/10.1111/1468-0262.00153.}


\bibitem[\protect\citeauthoryear{Heilmeier, Wischnewski, Hermansdorfer, Betz, Lienkamp, and Lohmann}{Heilmeier et~al\mbox{.}}{2020}]%
        {heilmeier_minimum_2020}
\bibfield{author}{\bibinfo{person}{Alexander Heilmeier}, \bibinfo{person}{Alexander Wischnewski}, \bibinfo{person}{Leonhard Hermansdorfer}, \bibinfo{person}{Johannes Betz}, \bibinfo{person}{Markus Lienkamp}, {and} \bibinfo{person}{Boris Lohmann}.} \bibinfo{year}{2020}\natexlab{}.
\newblock \showarticletitle{Minimum curvature trajectory planning and control for an autonomous race car}.
\newblock \bibinfo{journal}{\emph{Vehicle System Dynamics}} \bibinfo{volume}{58}, \bibinfo{number}{10} (\bibinfo{date}{Oct.} \bibinfo{year}{2020}), \bibinfo{pages}{1497--1527}.
\newblock
\showISSN{0042-3114}
\urldef\tempurl%
\url{https://doi.org/10.1080/00423114.2019.1631455}
\showDOI{\tempurl}
\newblock
\shownote{Publisher: Taylor \& Francis.}


\bibitem[\protect\citeauthoryear{Howard}{Howard}{2009}]%
        {howard_adaptive_2009}
\bibfield{author}{\bibinfo{person}{Thomas~M Howard}.} \bibinfo{year}{2009}\natexlab{}.
\newblock \bibinfo{booktitle}{\emph{Adaptive model-predictive motion planning for navigation in complex environments}}.
\newblock \bibinfo{publisher}{Carnegie Mellon University}.
\newblock


\bibitem[\protect\citeauthoryear{Howard and Kelly}{Howard and Kelly}{2007}]%
        {howard_optimal_2007}
\bibfield{author}{\bibinfo{person}{Thomas~M. Howard} {and} \bibinfo{person}{Alonzo Kelly}.} \bibinfo{year}{2007}\natexlab{}.
\newblock \showarticletitle{Optimal {Rough} {Terrain} {Trajectory} {Generation} for {Wheeled} {Mobile} {Robots}}.
\newblock \bibinfo{journal}{\emph{The International Journal of Robotics Research}} \bibinfo{volume}{26}, \bibinfo{number}{2} (\bibinfo{date}{Feb.} \bibinfo{year}{2007}), \bibinfo{pages}{141--166}.
\newblock
\showISSN{0278-3649}
\urldef\tempurl%
\url{https://doi.org/10.1177/0278364906075328}
\showDOI{\tempurl}
\newblock
\shownote{Publisher: SAGE Publications Ltd STM.}


\bibitem[\protect\citeauthoryear{Huang, Dossa, Ye, Braga, Chakraborty, Mehta, and Araújo}{Huang et~al\mbox{.}}{2022}]%
        {huang2022cleanrl}
\bibfield{author}{\bibinfo{person}{Shengyi Huang}, \bibinfo{person}{Rousslan Fernand~Julien Dossa}, \bibinfo{person}{Chang Ye}, \bibinfo{person}{Jeff Braga}, \bibinfo{person}{Dipam Chakraborty}, \bibinfo{person}{Kinal Mehta}, {and} \bibinfo{person}{João~G.M. Araújo}.} \bibinfo{year}{2022}\natexlab{}.
\newblock \showarticletitle{CleanRL: High-quality Single-file Implementations of Deep Reinforcement Learning Algorithms}.
\newblock \bibinfo{journal}{\emph{Journal of Machine Learning Research}} \bibinfo{volume}{23}, \bibinfo{number}{274} (\bibinfo{year}{2022}), \bibinfo{pages}{1--18}.
\newblock
\urldef\tempurl%
\url{http://jmlr.org/papers/v23/21-1342.html}
\showURL{%
\tempurl}


\bibitem[\protect\citeauthoryear{Huang, Zambelli, Kay, Martins, Tassa, Pilarski, and Hadsell}{Huang et~al\mbox{.}}{2019}]%
        {huang_learning_2019}
\bibfield{author}{\bibinfo{person}{Sandy~H. Huang}, \bibinfo{person}{Martina Zambelli}, \bibinfo{person}{Jackie Kay}, \bibinfo{person}{Murilo~F. Martins}, \bibinfo{person}{Yuval Tassa}, \bibinfo{person}{Patrick~M. Pilarski}, {and} \bibinfo{person}{Raia Hadsell}.} \bibinfo{year}{2019}\natexlab{}.
\newblock \bibinfo{title}{Learning {Gentle} {Object} {Manipulation} with {Curiosity}-{Driven} {Deep} {Reinforcement} {Learning}}.
\newblock
\newblock
\urldef\tempurl%
\url{https://doi.org/10.48550/arXiv.1903.08542}
\showDOI{\tempurl}
\newblock
\shownote{arXiv:1903.08542 [cs].}


\bibitem[\protect\citeauthoryear{Jin, Keutzer, and Levine}{Jin et~al\mbox{.}}{2018}]%
        {jin_regret_2018}
\bibfield{author}{\bibinfo{person}{Peter Jin}, \bibinfo{person}{Kurt Keutzer}, {and} \bibinfo{person}{Sergey Levine}.} \bibinfo{year}{2018}\natexlab{}.
\newblock \bibinfo{title}{Regret {Minimization} for {Partially} {Observable} {Deep} {Reinforcement} {Learning}}.
\newblock
\newblock
\urldef\tempurl%
\url{http://arxiv.org/abs/1710.11424}
\showURL{%
\tempurl}
\newblock
\shownote{arXiv:1710.11424 [cs].}


\bibitem[\protect\citeauthoryear{Kaiser, Babaeizadeh, Milos, Osinski, Campbell, Czechowski, Erhan, Finn, Kozakowski, Levine, Mohiuddin, Sepassi, Tucker, and Michalewski}{Kaiser et~al\mbox{.}}{2020}]%
        {kaiser_model-based_2020}
\bibfield{author}{\bibinfo{person}{Lukasz Kaiser}, \bibinfo{person}{Mohammad Babaeizadeh}, \bibinfo{person}{Piotr Milos}, \bibinfo{person}{Blazej Osinski}, \bibinfo{person}{Roy~H. Campbell}, \bibinfo{person}{Konrad Czechowski}, \bibinfo{person}{Dumitru Erhan}, \bibinfo{person}{Chelsea Finn}, \bibinfo{person}{Piotr Kozakowski}, \bibinfo{person}{Sergey Levine}, \bibinfo{person}{Afroz Mohiuddin}, \bibinfo{person}{Ryan Sepassi}, \bibinfo{person}{George Tucker}, {and} \bibinfo{person}{Henryk Michalewski}.} \bibinfo{year}{2020}\natexlab{}.
\newblock \bibinfo{title}{Model-{Based} {Reinforcement} {Learning} for {Atari}}.
\newblock
\newblock
\urldef\tempurl%
\url{https://doi.org/10.48550/arXiv.1903.00374}
\showDOI{\tempurl}
\newblock
\shownote{arXiv:1903.00374 [cs, stat].}


\bibitem[\protect\citeauthoryear{Kelly and Nagy}{Kelly and Nagy}{2003}]%
        {kelly_reactive_2003}
\bibfield{author}{\bibinfo{person}{Alonzo Kelly} {and} \bibinfo{person}{Bryan Nagy}.} \bibinfo{year}{2003}\natexlab{}.
\newblock \showarticletitle{Reactive {Nonholonomic} {Trajectory} {Generation} via {Parametric} {Optimal} {Control}}.
\newblock \bibinfo{journal}{\emph{The International Journal of Robotics Research}} \bibinfo{volume}{22}, \bibinfo{number}{7-8} (\bibinfo{date}{July} \bibinfo{year}{2003}), \bibinfo{pages}{583--601}.
\newblock
\showISSN{0278-3649}
\urldef\tempurl%
\url{https://doi.org/10.1177/02783649030227008}
\showDOI{\tempurl}
\newblock
\shownote{Publisher: SAGE Publications Ltd STM.}


\bibitem[\protect\citeauthoryear{Kulesza and Taskar}{Kulesza and Taskar}{2012}]%
        {kulesza_determinantal_2012}
\bibfield{author}{\bibinfo{person}{Alex Kulesza} {and} \bibinfo{person}{Ben Taskar}.} \bibinfo{year}{2012}\natexlab{}.
\newblock \showarticletitle{Determinantal {Point} {Processes} for {Machine} {Learning}}.
\newblock \bibinfo{journal}{\emph{Foundations and Trends® in Machine Learning}} \bibinfo{volume}{5}, \bibinfo{number}{2–3} (\bibinfo{date}{Dec.} \bibinfo{year}{2012}), \bibinfo{pages}{123--286}.
\newblock
\showISSN{1935-8237, 1935-8245}
\urldef\tempurl%
\url{https://doi.org/10.1561/2200000044}
\showDOI{\tempurl}
\newblock
\shownote{Publisher: Now Publishers, Inc.}


\bibitem[\protect\citeauthoryear{Marler and Arora}{Marler and Arora}{2004}]%
        {marler_survey_2004}
\bibfield{author}{\bibinfo{person}{R.T. Marler} {and} \bibinfo{person}{J.S. Arora}.} \bibinfo{year}{2004}\natexlab{}.
\newblock \showarticletitle{Survey of multi-objective optimization methods for engineering}.
\newblock \bibinfo{journal}{\emph{Structural and Multidisciplinary Optimization}} \bibinfo{volume}{26}, \bibinfo{number}{6} (\bibinfo{date}{April} \bibinfo{year}{2004}), \bibinfo{pages}{369--395}.
\newblock
\showISSN{1615-1488}
\urldef\tempurl%
\url{https://doi.org/10.1007/s00158-003-0368-6}
\showDOI{\tempurl}


\bibitem[\protect\citeauthoryear{McNaughton, Urmson, Dolan, and Lee}{McNaughton et~al\mbox{.}}{2011}]%
        {mcnaughton_motion_2011}
\bibfield{author}{\bibinfo{person}{Matthew McNaughton}, \bibinfo{person}{Chris Urmson}, \bibinfo{person}{John~M. Dolan}, {and} \bibinfo{person}{Jin-Woo Lee}.} \bibinfo{year}{2011}\natexlab{}.
\newblock \showarticletitle{Motion planning for autonomous driving with a conformal spatiotemporal lattice}. In \bibinfo{booktitle}{\emph{2011 {IEEE} {International} {Conference} on {Robotics} and {Automation}}}. \bibinfo{pages}{4889--4895}.
\newblock
\urldef\tempurl%
\url{https://doi.org/10.1109/ICRA.2011.5980223}
\showDOI{\tempurl}
\newblock
\shownote{ISSN: 1050-4729.}


\bibitem[\protect\citeauthoryear{Mnih, Kavukcuoglu, Silver, Graves, Antonoglou, Wierstra, and Riedmiller}{Mnih et~al\mbox{.}}{2013}]%
        {mnih_playing_2013}
\bibfield{author}{\bibinfo{person}{Volodymyr Mnih}, \bibinfo{person}{Koray Kavukcuoglu}, \bibinfo{person}{David Silver}, \bibinfo{person}{Alex Graves}, \bibinfo{person}{Ioannis Antonoglou}, \bibinfo{person}{Daan Wierstra}, {and} \bibinfo{person}{Martin Riedmiller}.} \bibinfo{year}{2013}\natexlab{}.
\newblock \bibinfo{title}{Playing {Atari} with {Deep} {Reinforcement} {Learning}}.
\newblock
\newblock
\urldef\tempurl%
\url{https://doi.org/10.48550/arXiv.1312.5602}
\showDOI{\tempurl}
\newblock
\shownote{arXiv:1312.5602 [cs].}


\bibitem[\protect\citeauthoryear{Nagy and Kelly}{Nagy and Kelly}{[n.d.]}]%
        {nagy_trajectory_nodate}
\bibfield{author}{\bibinfo{person}{Bryan Nagy} {and} \bibinfo{person}{Alonzo Kelly}.} \bibinfo{year}{[n.d.]}\natexlab{}.
\newblock \showarticletitle{{TRAJECTORY} {GENERATION} {FOR} {CAR}-{LIKE} {ROBOTS} {USING} {CUBIC} {CURVATURE} {POLYNOMIALS}}.
\newblock  (\bibinfo{year}{[n.\,d.]}).
\newblock


\bibitem[\protect\citeauthoryear{Novati and Koumoutsakos}{Novati and Koumoutsakos}{2019}]%
        {novati_remember_2019}
\bibfield{author}{\bibinfo{person}{Guido Novati} {and} \bibinfo{person}{Petros Koumoutsakos}.} \bibinfo{year}{2019}\natexlab{}.
\newblock \bibinfo{title}{Remember and {Forget} for {Experience} {Replay}}.
\newblock
\newblock
\urldef\tempurl%
\url{https://doi.org/10.48550/arXiv.1807.05827}
\showDOI{\tempurl}
\newblock
\shownote{arXiv:1807.05827 [cs, stat].}


\bibitem[\protect\citeauthoryear{O'Kelly, Zheng, Karthik, and Mangharam}{O'Kelly et~al\mbox{.}}{2020}]%
        {okelly_f1tenth_2020}
\bibfield{author}{\bibinfo{person}{Matthew O'Kelly}, \bibinfo{person}{Hongrui Zheng}, \bibinfo{person}{Dhruv Karthik}, {and} \bibinfo{person}{Rahul Mangharam}.} \bibinfo{year}{2020}\natexlab{}.
\newblock \showarticletitle{{F1TENTH}: {An} {Open}-source {Evaluation} {Environment} for {Continuous} {Control} and {Reinforcement} {Learning}}.
\newblock \bibinfo{journal}{\emph{Proceedings of Machine Learning Research}}  \bibinfo{volume}{123} (\bibinfo{date}{April} \bibinfo{year}{2020}).
\newblock
\urldef\tempurl%
\url{https://par.nsf.gov/biblio/10221872-f1tenth-open-source-evaluation-environment-continuous-control-reinforcement-learning}
\showURL{%
\tempurl}


\bibitem[\protect\citeauthoryear{Schmidhuber}{Schmidhuber}{1990}]%
        {schmidhuber_-line_1990}
\bibfield{author}{\bibinfo{person}{J. Schmidhuber}.} \bibinfo{year}{1990}\natexlab{}.
\newblock \showarticletitle{An on-line algorithm for dynamic reinforcement learning and planning in reactive environments}. In \bibinfo{booktitle}{\emph{1990 {IJCNN} {International} {Joint} {Conference} on {Neural} {Networks}}}. \bibinfo{publisher}{IEEE}, \bibinfo{address}{San Diego, CA, USA}, \bibinfo{pages}{253--258 vol.2}.
\newblock
\urldef\tempurl%
\url{https://doi.org/10.1109/IJCNN.1990.137723}
\showDOI{\tempurl}


\bibitem[\protect\citeauthoryear{Schulman, Wolski, Dhariwal, Radford, and Klimov}{Schulman et~al\mbox{.}}{2017}]%
        {schulman_proximal_2017}
\bibfield{author}{\bibinfo{person}{John Schulman}, \bibinfo{person}{Filip Wolski}, \bibinfo{person}{Prafulla Dhariwal}, \bibinfo{person}{Alec Radford}, {and} \bibinfo{person}{Oleg Klimov}.} \bibinfo{year}{2017}\natexlab{}.
\newblock \bibinfo{title}{Proximal {Policy} {Optimization} {Algorithms}}.
\newblock
\newblock
\urldef\tempurl%
\url{https://doi.org/10.48550/arXiv.1707.06347}
\showDOI{\tempurl}
\newblock
\shownote{arXiv:1707.06347 [cs].}


\bibitem[\protect\citeauthoryear{Schwarting, Seyde, Gilitschenski, Liebenwein, Sander, Karaman, and Rus}{Schwarting et~al\mbox{.}}{2021}]%
        {schwarting_deep_2021}
\bibfield{author}{\bibinfo{person}{Wilko Schwarting}, \bibinfo{person}{Tim Seyde}, \bibinfo{person}{Igor Gilitschenski}, \bibinfo{person}{Lucas Liebenwein}, \bibinfo{person}{Ryan Sander}, \bibinfo{person}{Sertac Karaman}, {and} \bibinfo{person}{Daniela Rus}.} \bibinfo{year}{2021}\natexlab{}.
\newblock \bibinfo{title}{Deep {Latent} {Competition}: {Learning} to {Race} {Using} {Visual} {Control} {Policies} in {Latent} {Space}}.
\newblock
\newblock
\urldef\tempurl%
\url{https://doi.org/10.48550/arXiv.2102.09812}
\showDOI{\tempurl}
\newblock
\shownote{arXiv:2102.09812 [cs].}


\bibitem[\protect\citeauthoryear{Seyde, Gilitschenski, Schwarting, Stellato, Riedmiller, Wulfmeier, and Rus}{Seyde et~al\mbox{.}}{2021}]%
        {seyde_is_2021}
\bibfield{author}{\bibinfo{person}{Tim Seyde}, \bibinfo{person}{Igor Gilitschenski}, \bibinfo{person}{Wilko Schwarting}, \bibinfo{person}{Bartolomeo Stellato}, \bibinfo{person}{Martin Riedmiller}, \bibinfo{person}{Markus Wulfmeier}, {and} \bibinfo{person}{Daniela Rus}.} \bibinfo{year}{2021}\natexlab{}.
\newblock \bibinfo{title}{Is {Bang}-{Bang} {Control} {All} {You} {Need}? {Solving} {Continuous} {Control} with {Bernoulli} {Policies}}.
\newblock
\newblock
\urldef\tempurl%
\url{https://doi.org/10.48550/arXiv.2111.02552}
\showDOI{\tempurl}
\newblock
\shownote{arXiv:2111.02552 [cs].}


\bibitem[\protect\citeauthoryear{Sinha, O’Kelly, Zheng, Mangharam, Duchi, and Tedrake}{Sinha et~al\mbox{.}}{2020}]%
        {sinha_formulazero_2020}
\bibfield{author}{\bibinfo{person}{Aman Sinha}, \bibinfo{person}{Matthew O’Kelly}, \bibinfo{person}{Hongrui Zheng}, \bibinfo{person}{Rahul Mangharam}, \bibinfo{person}{John Duchi}, {and} \bibinfo{person}{Russ Tedrake}.} \bibinfo{year}{2020}\natexlab{}.
\newblock \showarticletitle{{FormulaZero}: {Distributionally} {Robust} {Online} {Adaptation} via {Offline} {Population} {Synthesis}}. In \bibinfo{booktitle}{\emph{Proceedings of the 37th {International} {Conference} on {Machine} {Learning}}}. \bibinfo{publisher}{PMLR}, \bibinfo{pages}{8992--9004}.
\newblock
\urldef\tempurl%
\url{https://proceedings.mlr.press/v119/sinha20a.html}
\showURL{%
\tempurl}
\newblock
\shownote{ISSN: 2640-3498.}


\bibitem[\protect\citeauthoryear{Sunehag, Lever, Gruslys, Czarnecki, Zambaldi, Jaderberg, Lanctot, Sonnerat, Leibo, Tuyls, and Graepel}{Sunehag et~al\mbox{.}}{2017}]%
        {sunehag_value-decomposition_2017}
\bibfield{author}{\bibinfo{person}{Peter Sunehag}, \bibinfo{person}{Guy Lever}, \bibinfo{person}{Audrunas Gruslys}, \bibinfo{person}{Wojciech~Marian Czarnecki}, \bibinfo{person}{Vinicius Zambaldi}, \bibinfo{person}{Max Jaderberg}, \bibinfo{person}{Marc Lanctot}, \bibinfo{person}{Nicolas Sonnerat}, \bibinfo{person}{Joel~Z. Leibo}, \bibinfo{person}{Karl Tuyls}, {and} \bibinfo{person}{Thore Graepel}.} \bibinfo{year}{2017}\natexlab{}.
\newblock \bibinfo{title}{Value-{Decomposition} {Networks} {For} {Cooperative} {Multi}-{Agent} {Learning}}.
\newblock
\newblock
\urldef\tempurl%
\url{http://arxiv.org/abs/1706.05296}
\showURL{%
\tempurl}
\newblock
\shownote{arXiv:1706.05296 [cs].}


\bibitem[\protect\citeauthoryear{Sutton and Barto}{Sutton and Barto}{2018}]%
        {sutton_reinforcement_2018}
\bibfield{author}{\bibinfo{person}{Richard~S. Sutton} {and} \bibinfo{person}{Andrew~G. Barto}.} \bibinfo{year}{2018}\natexlab{}.
\newblock \bibinfo{booktitle}{\emph{Reinforcement {Learning}, second edition: {An} {Introduction}}}.
\newblock \bibinfo{publisher}{MIT Press}.
\newblock
\showISBNx{978-0-262-35270-3}
\newblock
\shownote{Google-Books-ID: uWV0DwAAQBAJ.}


\bibitem[\protect\citeauthoryear{Thuruthel, Falotico, Renda, and Laschi}{Thuruthel et~al\mbox{.}}{2019}]%
        {thuruthel_model-based_2019}
\bibfield{author}{\bibinfo{person}{Thomas~George Thuruthel}, \bibinfo{person}{Egidio Falotico}, \bibinfo{person}{Federico Renda}, {and} \bibinfo{person}{Cecilia Laschi}.} \bibinfo{year}{2019}\natexlab{}.
\newblock \showarticletitle{Model-{Based} {Reinforcement} {Learning} for {Closed}-{Loop} {Dynamic} {Control} of {Soft} {Robotic} {Manipulators}}.
\newblock \bibinfo{journal}{\emph{IEEE Transactions on Robotics}} \bibinfo{volume}{35}, \bibinfo{number}{1} (\bibinfo{date}{Feb.} \bibinfo{year}{2019}), \bibinfo{pages}{124--134}.
\newblock
\showISSN{1552-3098, 1941-0468}
\urldef\tempurl%
\url{https://doi.org/10.1109/TRO.2018.2878318}
\showDOI{\tempurl}


\bibitem[\protect\citeauthoryear{Werling, Ziegler, Kammel, and Thrun}{Werling et~al\mbox{.}}{2010}]%
        {werling2010optimal}
\bibfield{author}{\bibinfo{person}{Moritz Werling}, \bibinfo{person}{Julius Ziegler}, \bibinfo{person}{S{\"o}ren Kammel}, {and} \bibinfo{person}{Sebastian Thrun}.} \bibinfo{year}{2010}\natexlab{}.
\newblock \showarticletitle{Optimal trajectory generation for dynamic street scenarios in a frenet frame}. In \bibinfo{booktitle}{\emph{2010 IEEE international conference on robotics and automation}}. IEEE, \bibinfo{pages}{987--993}.
\newblock


\bibitem[\protect\citeauthoryear{Xie, Losey, Tolsma, Finn, and Sadigh}{Xie et~al\mbox{.}}{2021}]%
        {xie_learning_2021}
\bibfield{author}{\bibinfo{person}{Annie Xie}, \bibinfo{person}{Dylan Losey}, \bibinfo{person}{Ryan Tolsma}, \bibinfo{person}{Chelsea Finn}, {and} \bibinfo{person}{Dorsa Sadigh}.} \bibinfo{year}{2021}\natexlab{}.
\newblock \showarticletitle{Learning {Latent} {Representations} to {Influence} {Multi}-{Agent} {Interaction}}. In \bibinfo{booktitle}{\emph{Proceedings of the 2020 {Conference} on {Robot} {Learning}}}. \bibinfo{publisher}{PMLR}, \bibinfo{pages}{575--588}.
\newblock
\urldef\tempurl%
\url{https://proceedings.mlr.press/v155/xie21a.html}
\showURL{%
\tempurl}
\newblock
\shownote{ISSN: 2640-3498.}


\bibitem[\protect\citeauthoryear{Zhang, Vikram, Smith, Abbeel, Johnson, and Levine}{Zhang et~al\mbox{.}}{2019}]%
        {zhang_solar_2019}
\bibfield{author}{\bibinfo{person}{Marvin Zhang}, \bibinfo{person}{Sharad Vikram}, \bibinfo{person}{Laura Smith}, \bibinfo{person}{Pieter Abbeel}, \bibinfo{person}{Matthew Johnson}, {and} \bibinfo{person}{Sergey Levine}.} \bibinfo{year}{2019}\natexlab{}.
\newblock \showarticletitle{{SOLAR}: {Deep} {Structured} {Representations} for {Model}-{Based} {Reinforcement} {Learning}}. In \bibinfo{booktitle}{\emph{Proceedings of the 36th {International} {Conference} on {Machine} {Learning}}}. \bibinfo{publisher}{PMLR}, \bibinfo{pages}{7444--7453}.
\newblock
\urldef\tempurl%
\url{https://proceedings.mlr.press/v97/zhang19m.html}
\showURL{%
\tempurl}
\newblock
\shownote{ISSN: 2640-3498.}


\bibitem[\protect\citeauthoryear{Zinkevich, Johanson, Bowling, and Piccione}{Zinkevich et~al\mbox{.}}{2007}]%
        {zinkevich_regret_2007}
\bibfield{author}{\bibinfo{person}{Martin Zinkevich}, \bibinfo{person}{Michael Johanson}, \bibinfo{person}{Michael Bowling}, {and} \bibinfo{person}{Carmelo Piccione}.} \bibinfo{year}{2007}\natexlab{}.
\newblock \showarticletitle{Regret {Minimization} in {Games} with {Incomplete} {Information}}. In \bibinfo{booktitle}{\emph{Advances in {Neural} {Information} {Processing} {Systems}}}, Vol.~\bibinfo{volume}{20}. \bibinfo{publisher}{Curran Associates, Inc.}
\newblock
\urldef\tempurl%
\url{https://proceedings.neurips.cc/paper/2007/hash/08d98638c6fcd194a4b1e6992063e944-Abstract.html}
\showURL{%
\tempurl}


\end{thebibliography}
